\documentclass{article}

\usepackage[preprint]{neurips_2023}
\usepackage[utf8]{inputenc}
\usepackage[T1]{fontenc}
\usepackage{hyperref}
\usepackage{url}
\usepackage{booktabs}
\usepackage{amsfonts}
\usepackage{amsmath,amssymb}
\usepackage{graphicx}
\usepackage{microtype}
\usepackage{xcolor}
\usepackage{tabularx}
\usepackage{caption}
\usepackage{siunitx}
\usepackage{enumitem}
\usepackage{array}
\usepackage[capitalize,noabbrev]{cleveref}

\makeatletter
\renewcommand{\@notice}{}
\makeatother

\title{Where Does Texture Evidence Live in SAM?\\Features, Proposal Masks, and Texture Segmentation}
\author{%
  Nadav Orenstein \\
  Tel Aviv University, Israel \\
  \texttt{nadavo1@mail.tau.ac.il} \\
  \And
  Aviad Cohen Zada \\
  Tel Aviv University, Israel \\
  \texttt{cohenzada@mail.tau.ac.il} \\
  \And
  Shai Avidan \\
  Tel Aviv University, Israel \\
  \texttt{avidan@eng.tau.ac.il} \\
  \And
  Gal Oren \\
  Stanford University, Technion, USA \\
  \texttt{galoren@stanford.edu} \\
}

\begin{document}

\maketitle

\begin{abstract}
Texture segmentation is a semantic-sparse stress test for foundation segmentation: meaningful regions are defined by material or repeated appearance rather than by object identity. Segment Anything Models (SAMs) often fail by default on such texture-defined partitions, but that failure is ambiguous. It may reflect missing texture evidence, missing proposal-bank support, or a default object-centric readout that selects or assembles the wrong evidence. Prior texture-aware adaptation work shows that SAM-family models can be pushed toward better texture behavior through fine-tuning. We ask a different question: before adaptation, what texture-relevant evidence is already preserved in frozen SAM? We study two frozen evidence spaces. First, we probe frozen multiscale features with a minimal diagnostic clustering readout to test whether they retain enough organization to separate the target partition. Second, we treat the frozen automatic proposal bank as evidence rather than output, and apply a supervised consolidation readout to assemble texture-aligned masks or fragments into an end-task segmentation. In both cases SAM remains frozen: there is no backbone fine-tuning and no proposal-generator retraining. The goal is texture segmentation from preserved evidence, not texture reconstruction. Our benchmark design starts from the two historical anchors of texture segmentation: RWTD on natural images and the STLD synthetic lineage. We broaden the natural side with an ADE20K-selected refined-crop complement included in the supplementary package, and the synthetic side with a canonical ControlNet-stitched PTD evaluation archive containing 1,742 images and occupying about 800 MB after retention and packaging. Across these routes, frozen SAM is not a texture segmenter by default, but its failures are not simple texture blindness. Coarse frozen features preserve texture organization, and proposal banks often contain texture-aligned masks or fragments. The regimes differ: natural scenes more often require assembly and commitment over fragments, whereas cleaner synthetic cases more often reduce to selecting an already coherent proposal. Default mask failure should therefore be decomposed into preserved representation evidence, proposal-bank support, readout mismatch, and commitment failure. Code is available at https://github.com/Scientific-Computing-Lab/ArchiTexture
\end{abstract}

\section{Introduction}
\label{sec:intro}
Segment Anything Models have made high-recall foundation segmentation broadly available, but their interface and default behavior remain strongly object-centric~\cite{kirillov2023sam,ravi2024sam2}. This matters because many important segmentation targets are not naturally object-centric. Material transitions in everyday scenes, repeated cellular structure in microscopy, patterned land cover in remote sensing, and manufactured or woven surfaces all produce boundaries that are real but defined by material, microstructure, or repeated appearance rather than by object identity.

Texture-aware segmentation is therefore not simply semantic segmentation with unusual labels~\cite{cimpoi2014dtd,bell2015minc,mikes2022texturebenchmark}. In such settings, the target partition is governed by local structure, orientation, granularity, and weak transition cues rather than by named semantic categories. Standard semantic benchmarks often allow a model to exploit object category, scene context, or canonical shape; texture-defined partitions deliberately weaken those cues. Texture segmentation is therefore a semantic-sparse stress test for segmentation foundation models: success should come from appearance transitions and region-level texture organization rather than from recognizing familiar semantics.

SAM-style automatic outputs often fail on these texture-defined partitions, fragmenting coherent textures or committing to semantic shapes instead of appearance transitions. Scientifically, however, that failure is ambiguous. It may indicate that the representation lacks useful texture information, that the automatic proposal bank omits the needed support, or that the evidence is present but selected or assembled incorrectly by the default object-centric readout.

TextureSAM is the closest prior work in this setting because it shows that SAM-family models can be adapted toward stronger texture-aware behavior through controlled texture replacement and fine-tuning~\cite{cohen2025texturesam,cohen2026texturesam}. That line of work establishes that SAM's texture failure mode matters and that changing the model can improve it. We ask a complementary evaluative question: when SAM fails by default on a texture-defined partition, is the relevant texture information absent, or is it already preserved in frozen features and proposal masks but missed by the default readout?

\begin{figure*}[t]
\centering
\includegraphics[width=\textwidth]{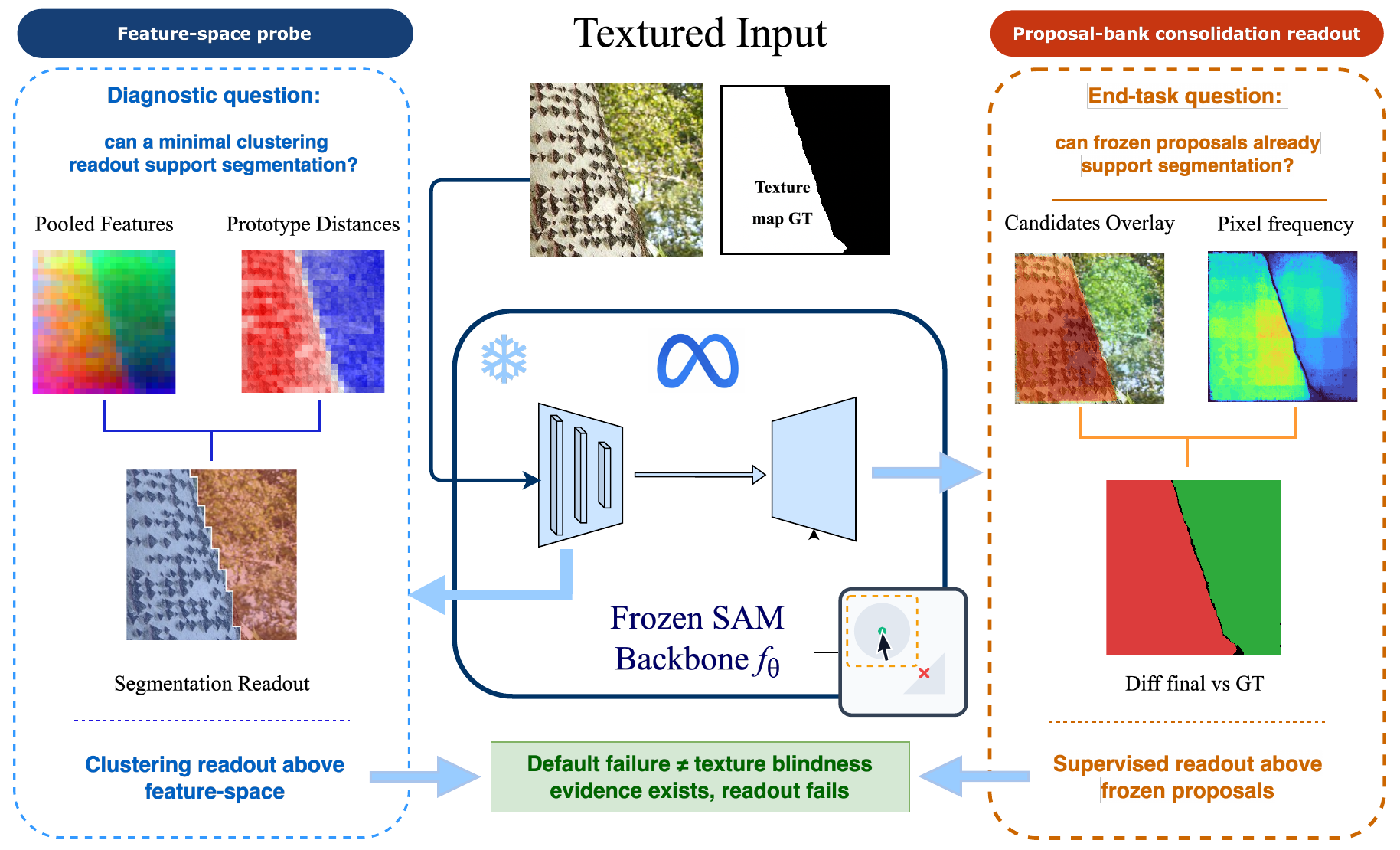}
\caption{
\textbf{Two complementary probes test whether texture-defined segmentation is recoverable from frozen SAM.}
The left branch probes frozen feature space, asking whether pooled multiscale features support a minimal clustering readout. The right branch probes the frozen automatic proposal bank, asking whether a consolidation readout can assemble proposal evidence into the target partition. Together, the two branches distinguish missing texture evidence from readout failure.
}
\label{fig:recoverability_loci}
\end{figure*}

We study two separate evidence spaces exposed by a frozen SAM-style segmenter: frozen multiscale features and the frozen automatic proposal bank. The feature-space route is diagnostic, asking whether a minimal readout can separate the target partition from frozen features. The proposal-space route is the matched end-task readout, asking whether a lightweight consolidation layer can assemble frozen masks or fragments into the desired partition without fine-tuning SAM or retraining the proposal generator. As summarized in \Cref{fig:recoverability_loci}, the two branches query the same frozen SAM source but distinguish internal representation evidence from evidence already externalized as masks. This separation lets default mask failure be decomposed into missing evidence, missing proposal support, readout mismatch, or commitment failure.

% \begin{figure*}[t]
% \centering
% \includegraphics[width=01.05\textwidth]{figures/ArchiTexture.drawio.svg}
% \caption{Overview of the evaluation. The center shows the shared texture-defined input, the shared target partition, and the same frozen SAM source used in both branches. The left branch is a diagnostic feature-space probe that asks whether frozen multiscale features preserve enough texture organization for a minimal readout to separate the target partition. The right branch is a supervised proposal-bank consolidation readout that asks whether the frozen automatic mask bank already contains evidence that can be assembled into the desired segmentation. SAM remains frozen throughout: there is no backbone fine-tuning and no proposal-generator retraining. \shai{Looks good}}
% \label{fig:recoverability_loci}
% \end{figure*}
% \begin{figure*}[t]
% \centering
% \includegraphics[width=\textwidth]{figures/ArchiTexture.drawio}
% \caption{
% \textbf{Two complementary probes test whether texture-defined segmentation is recoverable from frozen SAM.}
% The left branch probes frozen feature space, asking whether pooled multiscale features support a minimal clustering readout. The right branch probes the frozen automatic proposal bank, asking whether a consolidation readout can assemble proposal evidence into the target partition. Together, the two branches distinguish missing texture evidence from readout failure.
% }
% \label{fig:recoverability_loci}
% \end{figure*}

The benchmark design is deliberately comparative rather than dataset-centric. We keep two historical anchors: RWTD as the natural-image anchor and STLD as the synthetic anchor. We add two breadth complements: an ADE20K-selected refined-crop complement, defined by a fixed crop manifest and paired refined masks rather than TextureSAM predictions~\cite{detextureade20k2026}, and a canonical ControlNet-stitched PTD evaluation archive containing 1,742 images and occupying about 800 MB after retention and packaging. The four-route design supports an evaluation claim based on agreement across regimes, not on any one route alone.

Why introduce Controlnet Bridge at all? The ControlNet bridge is needed because STLD gives exact labels but stylized hard seams, while natural images give richer transitions without exact generative control. It preserves an inherited binary layout from an exact synthetic scaffold while producing more naturalistic transitions than classical mosaics. Thus, it tests whether synthetic-side conclusions survive under a harder bridge regime without replacing STLD or natural benchmarks.

We make four tightly connected contributions.
\begin{itemize}[leftmargin=1.2em]
\item An evaluation framework that separates two frozen evidence spaces in SAM: multiscale features and automatic proposal masks.
\item A diagnostic feature-space probe showing that frozen SAM features can retain segmentation-relevant texture organization even when default masks fail.
\item A proposal-bank consolidation readout showing that frozen automatic mask banks often already contain useful texture-aligned masks or fragments, and that failures can arise from readout mismatch and commitment rather than missing evidence alone.
\item A structured benchmark design over RWTD, STLD, an ADE20K-selected refined-crop complement, and a ControlNet bridge, with explicit evaluator discipline and claim boundaries.
\end{itemize}

The paper is therefore not a new foundation model, not a stand-alone dataset paper, and not a claim that frozen SAM already solves texture segmentation. Its claim is narrower: frozen SAM often preserves segmentation-relevant texture evidence, internally in features and externally in proposal masks, and default failure becomes interpretable once evidence, readout, and commitment are separated.

\section{Related Work}
\label{sec:related}

\subsection{Texture Segmentation and Texture Benchmarks}

Texture segmentation has long been studied as a problem distinct from semantic recognition, with both descriptor-driven and learned approaches emphasizing appearance statistics, repeated local structure, and boundary ambiguity rather than named categories~\cite{cimpoi2014dtd,bell2015minc,cimpoi2016deepfilterbanks,andrearczyk2017texturefcn,liu2019bowtocnn}. Two historical anchors matter most for this paper. On the natural-image side, we use the Real-World Texture Dataset route (RWTD) consolidated in the Prague texture segmentation benchmark work~\cite{mikes2022texturebenchmark}. On the synthetic side, we use the shape-tailored local descriptor benchmark lineage (STLD) introduced by Khan et al.~\cite{khan2015stld,khan2017c2f,khan2018learnedstld}. We retain these anchors because they expose complementary regimes: the natural route stresses weak transitions and disconnected same-texture regions, while the synthetic route offers exact canonical partitions under controlled geometry.

Those anchors are useful but not sufficient on their own. RWTD is small, and STLD's hard synthetic transitions are cleaner than many modern image statistics. We therefore add two complements with narrower roles. The natural complement is a concurrently released ADE20K-selected refined-crop artifact included in the supplementary package~\cite{detextureade20k2026}. It is defined by a fixed crop manifest and paired refined masks, not by TextureSAM predictions, and it broadens the natural-image side beyond RWTD. The synthetic complement is a canonical ControlNet-stitched PTD evaluation archive with 1,742 images and about 800 MB of retained data~\cite{zhang2023controlnet,rombach2022ldm}. Both are complements to the anchors, not stand-alone dataset contributions.

\subsection{SAM and adaptation}

SAM and SAM2 provide strong foundation-segmentation interfaces, especially through high-recall proposal generation and promptable object-centric behavior~\cite{kirillov2023sam,ravi2024sam2}. A large body of follow-up work adapts those models to new domains or to stronger mask quality through prompt tuning, prompt learning, or decoder-side modification~\cite{ke2023hqsam,zhang2023persam,li2024asam}. AutoSAM is a representative example: it keeps the SAM backbone frozen but trains an auxiliary prompt encoder under supervised segmentation losses rather than reasoning over a frozen proposal bank~\cite{shaharabany2023autosam}.

TextureSAM is the closest prior work for our setting because it explicitly adapts SAM-family behavior for texture segmentation~\cite{cohen2025texturesam,cohen2026texturesam}. That line asks how SAM should be changed so texture behavior improves. We ask a different question: what evidence is already preserved before adaptation, and how should evaluation distinguish absent evidence from inaccessible or misread evidence? Instead of changing the backbone or retraining the proposal generator, we inspect what can be extracted from frozen SAM by constrained readouts.

This distinction also clarifies the role of ADE20K in our paper. TextureSAM is relevant there as an adapted comparator, but the ADE20K-selected complement itself is defined by the companion crop artifact rather than by TextureSAM predictions. Because the public TextureSAM checkpoint was trained on ADE20K, our comparison on that route is restricted to the validation slice. The next section turns these distinctions into explicit protocol boundaries.

\section{Texture Evidence in Frozen SAM}
\label{sec:problem}

\subsection{Retained Texture Information, Evidence Spaces, and Readouts}

Default texture failure is not a single failure mode. A frozen SAM-style segmenter may fail because the relevant partition is absent from its features, absent from its automatic proposal bank, or present but misread by the default object-centric readout. We therefore separate the frozen evidence SAM exposes from the readouts used to inspect it.

For an image \(x\), we inspect two frozen evidence spaces,
\(
\phi(x) \) and \( P(x),
\)
where \(\phi(x)\) denotes frozen multiscale features and \(P(x)\) denotes the automatic proposal bank. We apply two constrained readouts:
\[
\hat{Y}_{\mathrm{feat}} = R_{\mathrm{feat}}(\phi(x)), 
\qquad
\hat{Y}_{\mathrm{prop}} = R_{\mathrm{prop}}(P(x),x).
\]
SAM remains frozen in both cases: we do not fine-tune the backbone or retrain the proposal generator. A partition is recoverable when such a constrained readout can extract it from frozen SAM outputs. This tests segmentation-relevant texture evidence, not texture reconstruction, and does not imply that frozen SAM solves texture segmentation by default.

As \Cref{fig:recoverability_loci} summarizes, the two readouts have different roles. The feature-space route is diagnostic: a minimal clustering readout asks whether texture organization is latent in frozen features. The proposal-space route is the matched end-task readout: a lightweight consolidation layer asks whether frozen masks or fragments can be assembled into the desired partition. Together they distinguish missing evidence from readout mismatch or premature commitment to the wrong assembly.

\section{Feature-Space Probe}
\label{sec:feature}

\begin{figure*}[t]
    \centering
    \setlength{\tabcolsep}{1pt}
    \renewcommand{\arraystretch}{0.8}
    \begin{tabular}{@{}ccccc@{}}
        \textbf{Input} &
        \textbf{\begin{tabular}[c]{@{}c@{}}Pooled features \end{tabular}} &
        \textbf{\begin{tabular}[c]{@{}c@{}}Signed margin\end{tabular}} &
        \textbf{\begin{tabular}[c]{@{}c@{}}Pooled labels\end{tabular}} &
        \textbf{GT} \\
        \includegraphics[width=0.185\textwidth]{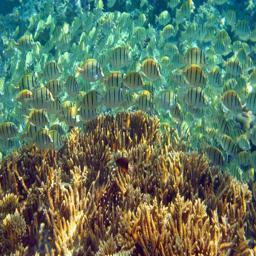} &
        \includegraphics[width=0.185\textwidth]{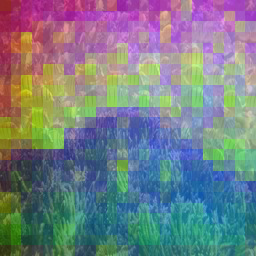} &
        \includegraphics[width=0.185\textwidth]{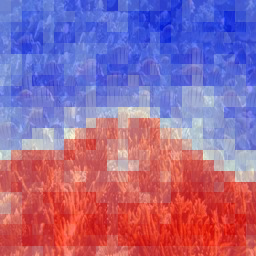} &
        \includegraphics[width=0.185\textwidth]{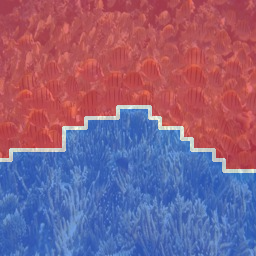} &
        \includegraphics[width=0.185\textwidth]{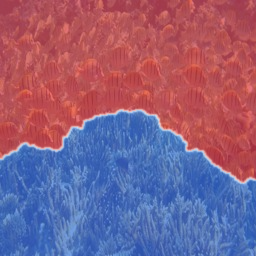} \\
        \includegraphics[width=0.185\textwidth]{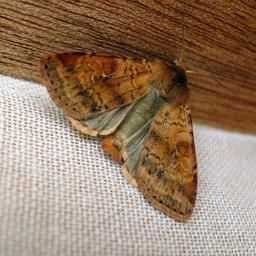} &
        \includegraphics[width=0.185\textwidth]{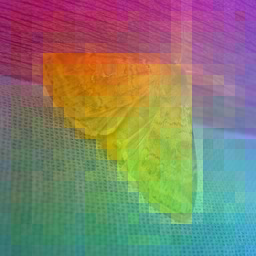} &
        \includegraphics[width=0.185\textwidth]{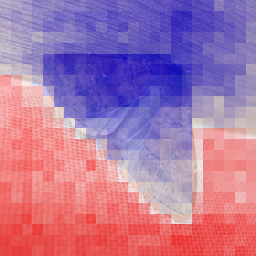} &
        \includegraphics[width=0.185\textwidth]{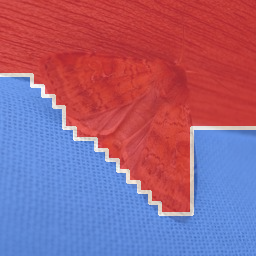} &
        \includegraphics[width=0.185\textwidth]{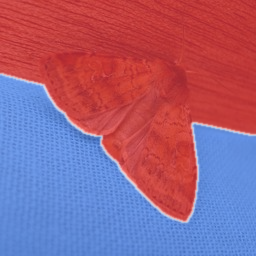} \\
        \includegraphics[width=0.185\textwidth]{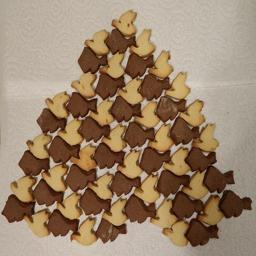} &
        \includegraphics[width=0.185\textwidth]{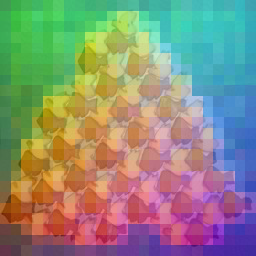} &
        \includegraphics[width=0.185\textwidth]{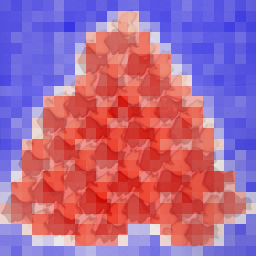} &
        \includegraphics[width=0.185\textwidth]{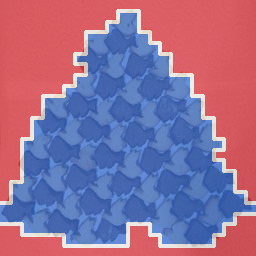} &
        \includegraphics[width=0.185\textwidth]{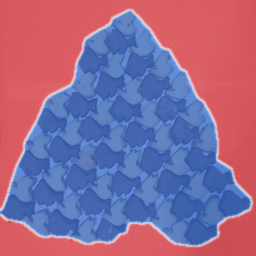}
    \end{tabular}
    \caption{
    Feature-space probe example on RWTD. From left to right:
    the original input image, the pooled SAM feature map visualized with PCA, signed proximity margin between the two fitted cluster means,
    pooled cluster-label prediction, ground-truth partition.
    All computation is performed in pooled feature space; relevant panels are upsampled and overlaid at image resolution for
    visualization.
    }    \label{fig:feature_probe}
\end{figure*}

\Cref{fig:feature_probe} expands the left-hand branch of \Cref{fig:recoverability_loci}. This section asks whether frozen SAM features preserve enough texture information for a minimal readout to separate the target partition. We use the probe as diagnostic evidence, not as the deployed benchmark readout: it never trains a decoder and never fine-tunes SAM. The feature-space row is included in the main results table as a diagnostic frozen-feature readout; it is interpreted separately from the proposal-space end-task readout because it uses a non-learned clustering protocol over frozen features, but it remains part of the reported evidence-space comparison.

Given an image \(x\), the probe extracts frozen multiscale SAM features, average-pools the coarsest feature map, clusters pooled spatial features into \(K=2\) groups, and upsamples the labels for visualization and scoring. Ground truth is used only for evaluation and label matching. Since all benchmark routes are binary texture-partition tasks, \(K=2\) is fixed rather than tuned per image. If this clustering-only readout meaningfully separates the benchmark sides, then frozen coarse features already contain texture-relevant structure, and default mask failure cannot be attributed to texture blindness alone.

The probe artifacts support that interpretation. Coarse pooled features are consistently more informative than finer native-grid controls. This suggests that the frozen representation carries structured but imperfect texture organization. The strongest use of this route is conceptual: repeated-pattern regions, gradual transitions, and several synthetic silhouette cases are already reflected in frozen feature space before any proposal reasoning or task-trained head is applied. Appendix~\ref{app:feature_provenance} gathers the parity table, qualitative gallery, and scale diagnostics behind this conclusion.

\section{Proposal-Bank Consolidation Readout}
\label{sec:proposal}

\begin{figure*}[t]
    \centering
    \includegraphics[width=0.7\textwidth]{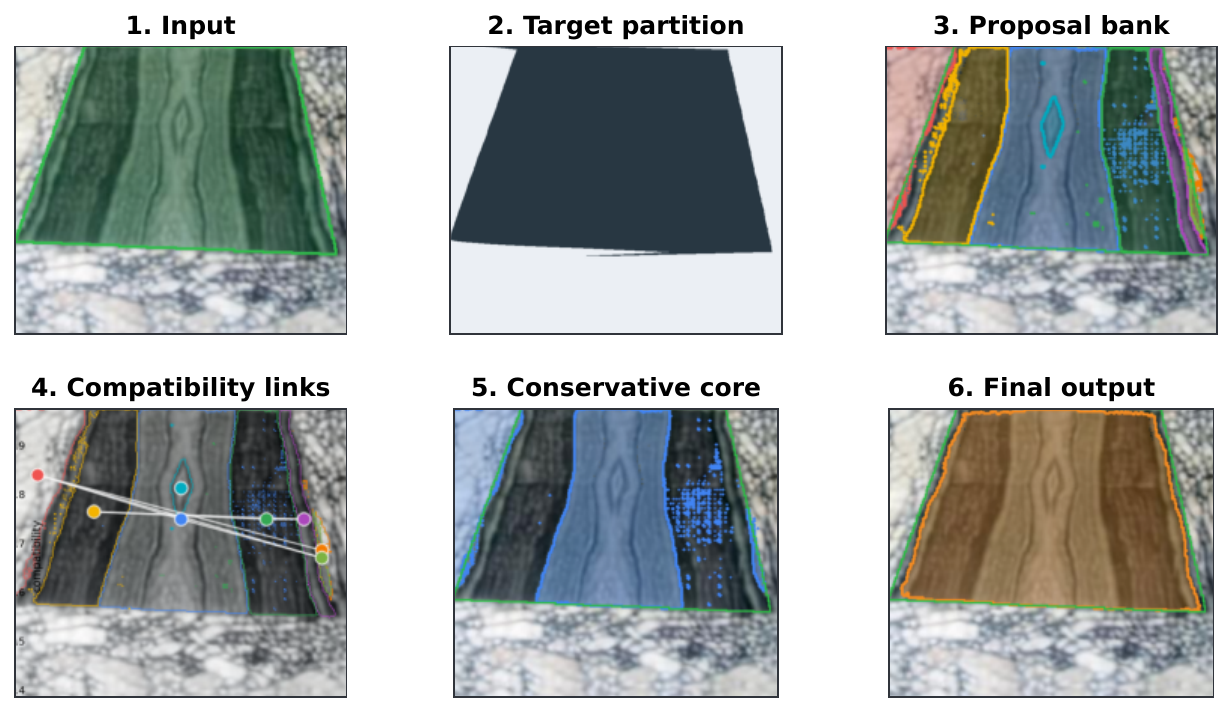}
    \caption{
    Proposal-bank consolidation above a frozen automatic mask bank. Frozen SAM emits candidate masks; the readout scores compatibility, forms candidate components, selects a conservative core, and optionally repairs it only when a learned policy predicts safe improvement. SAM itself is never fine-tuned.
    }
    \label{fig:proposal_logic}
\end{figure*}

\Cref{fig:proposal_logic} expands the right-hand branch of \Cref{fig:recoverability_loci}. This is the paper's matched end-task readout: after frozen SAM emits an automatic mask bank, can a constrained readout recover the desired texture partition from that bank? We keep the short name \emph{proposal-space recovery} in tables, but the component is technically a proposal-bank consolidation readout above frozen proposals. The generated masks are treated as evidence rather than as the final answer.

Concretely, the input is an image \(x\) and its frozen SAM automatic proposal bank
\(
\mathcal{M}(x)=\{m_i\}_{i=1}^{N},
\)
where each \(m_i\) is produced before any task-specific training. The output is a binary texture partition \(\hat{y}\). The algorithm does not modify SAM and does not generate new proposals. It learns a lightweight consolidation rule above the frozen bank: compute descriptors for each proposal, score nearby masks for texture-side compatibility, group compatible masks into candidate components, select a conservative low-overmerge core, and, on routes that use repair, switch to a denser candidate only when a learned safe-gain policy predicts improvement. Ground-truth partitions train the consolidation modules, but not the SAM backbone or proposal generator.

The supervision boundary is explicit. Proposal-space recovery is supervised in the technical sense: the compatibility model, component scorer, and optional repair policy are trained. However, the trained object is a consolidation rule above a frozen proposal bank. The main natural-image route does not use target-domain dense masks for training; instead, it uses generated partition supervision with exact binary labels by construction. Thus, if this readout extracts a strong texture partition, poor default masks do not imply absence of texture evidence in the bank. They indicate that the default object-centric selection rule is often misaligned with texture-defined partitioning.

Route-specific scope remains fixed. RWTD uses the full conservative-core plus optional-repair pipeline. The ADE20K-selected and ControlNet complements use the Stage-A proposal-space route without RWTD-specific repair. STLD uses synthetic layouts matched to its benchmark geometry, while the natural-image routes use generated partition supervision separated from target-domain dense masks. These distinctions are kept visible in the evaluation section and appendix rather than hidden behind one generic label.
\section{ControlNet Bridge Synthetic Complement}
\label{sec:controlnet_bridge}

STLD provides exact labels but uses stylized hard seams, while natural images provide richer transitions without exact control. We therefore add a synthetic bridge complement: each image starts from an exact two-region DTD stitch with inherited binary labels, then a ControlNet-conditioned Stable Diffusion v1.5 image-to-image pipeline re-renders the scaffold to make the boundary less trivial than a raw mosaic. The bridge generator uses a 500-step transition-tuned ControlNet checkpoint on a Stable Diffusion v1.5 base model; the generated bridge images are used only for evaluation, and no model is trained on those generated bridge images. This route complements STLD; it does not replace STLD or natural imagery.
For two sampled DTD textures~\cite{cimpoi2014dtd} and an exact mask \(M\), the scaffold is
\[
x_{\mathrm{stitch}} = M \odot x_A + (1-M)\odot x_B .
\]
The mask \(M\) is fixed before rendering and remains the evaluation label. We then forward-noise the stitched image to an intermediate timestep \(t'\),
\[
x_{t'} =
\sqrt{\bar{\alpha}_{t'}}\,x_{\mathrm{stitch}}
+
\sqrt{1-\bar{\alpha}_{t'}}\,\varepsilon,
\qquad
\varepsilon \sim \mathcal{N}(0,I),
\]
and denoise with a fixed prompt while conditioning on the scaffold-derived ControlNet signal~\cite{zhang2023controlnet,rombach2022ldm}. During bridge generation, the generative pipeline is fixed; only the rendered evaluation images vary through the sampled textures, masks, noise, and \(t'\).

\begin{figure*}[t]
    \centering
    \setlength{\tabcolsep}{2pt}
    \renewcommand{\arraystretch}{0.8}
    \begin{tabular}{@{}ccccc@{}}
        \textbf{region} & \textbf{hard-stitched} & \textbf{\(t'=367\)} & \textbf{\(t'=633\)} & \textbf{\(t'=900\)} \\
        \includegraphics[width=0.185\textwidth]{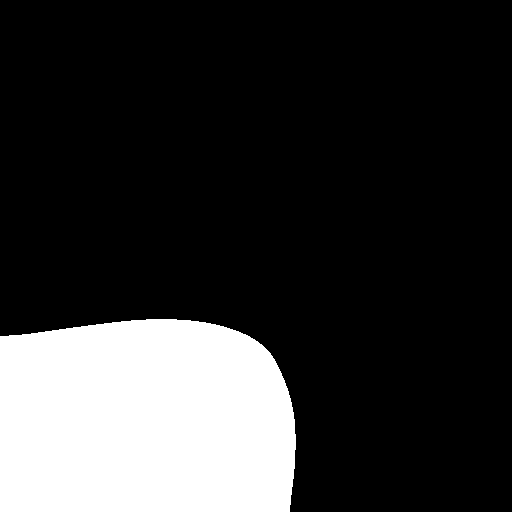} &
        \includegraphics[width=0.185\textwidth]{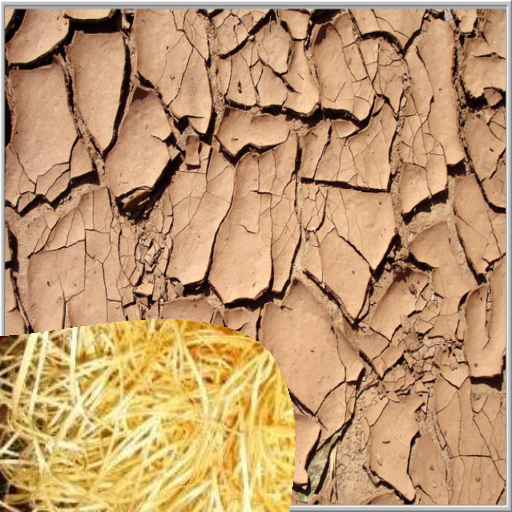} &
        \includegraphics[width=0.185\textwidth]{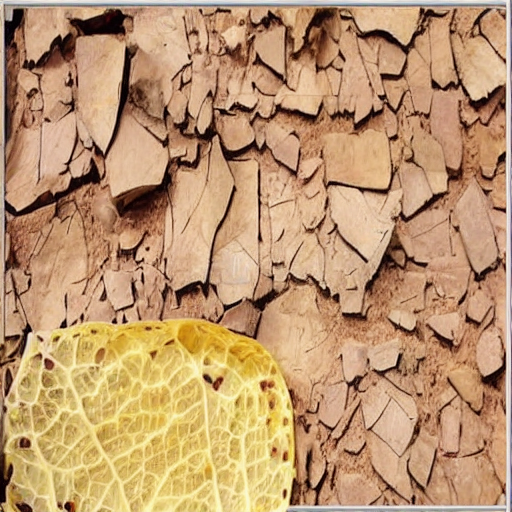} &
        \includegraphics[width=0.185\textwidth]{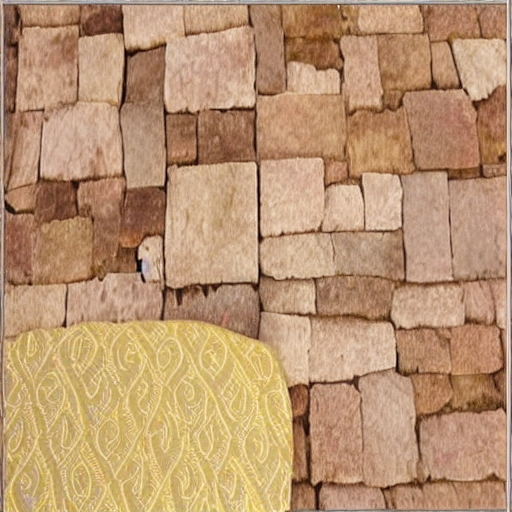} &
        \includegraphics[width=0.185\textwidth]{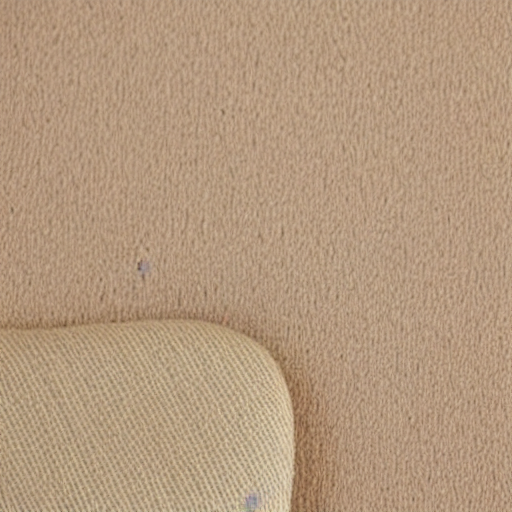} \\
        \includegraphics[width=0.185\textwidth]{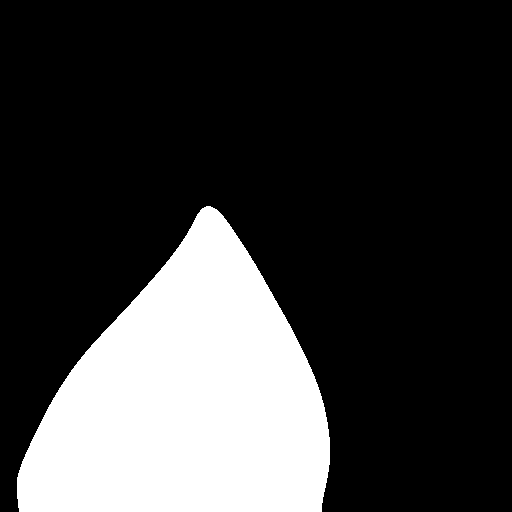} &
        \includegraphics[width=0.185\textwidth]{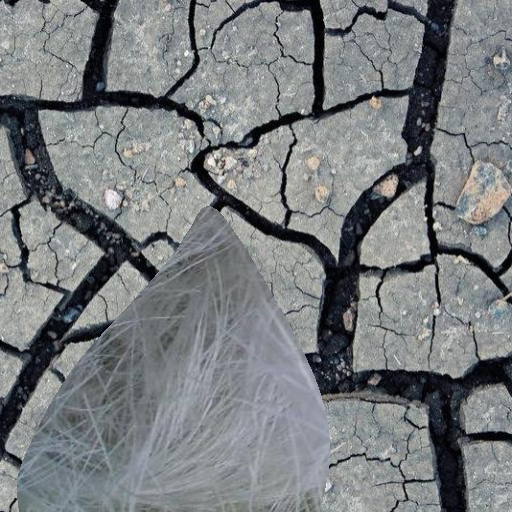} &
        \includegraphics[width=0.185\textwidth]{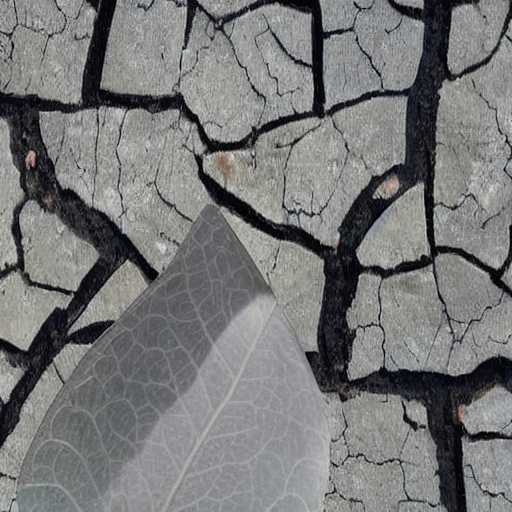} &
        \includegraphics[width=0.185\textwidth]{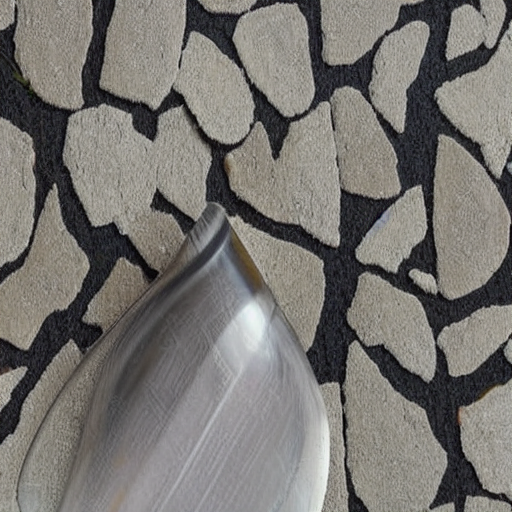} &
        \includegraphics[width=0.185\textwidth]{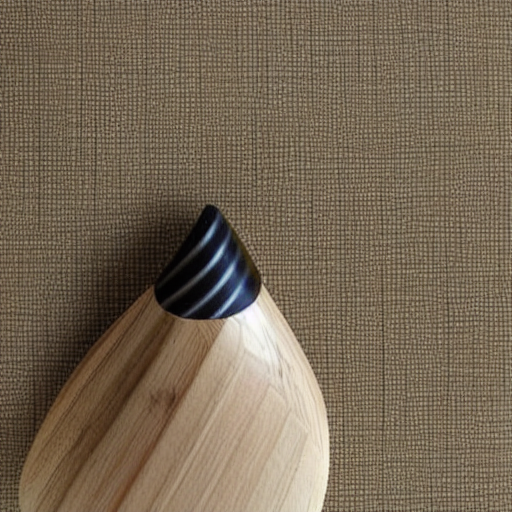} \\
        \includegraphics[width=0.185\textwidth]{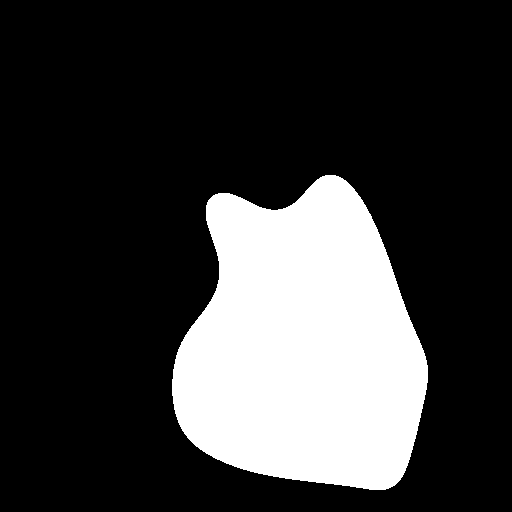} &
        \includegraphics[width=0.185\textwidth]{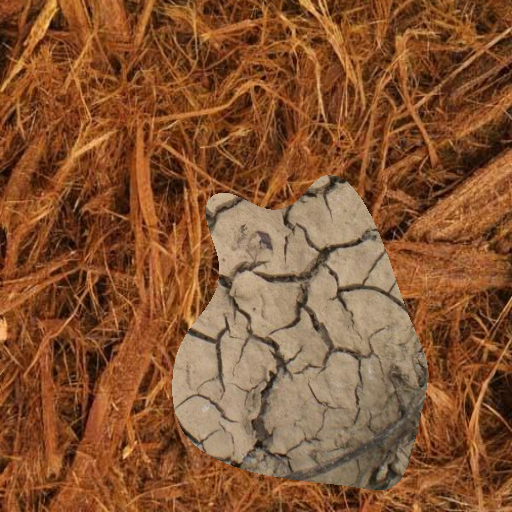} &
        \includegraphics[width=0.185\textwidth]{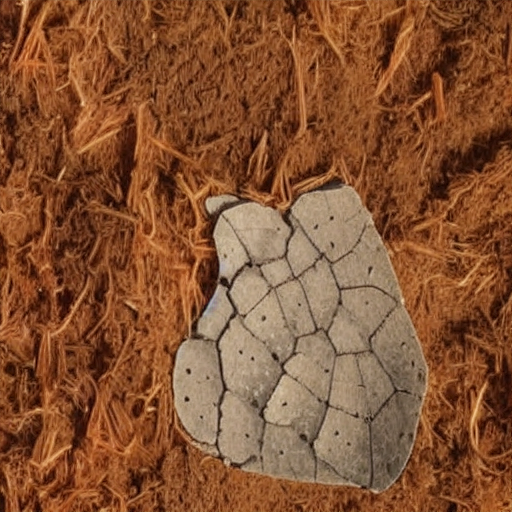} &
        \includegraphics[width=0.185\textwidth]{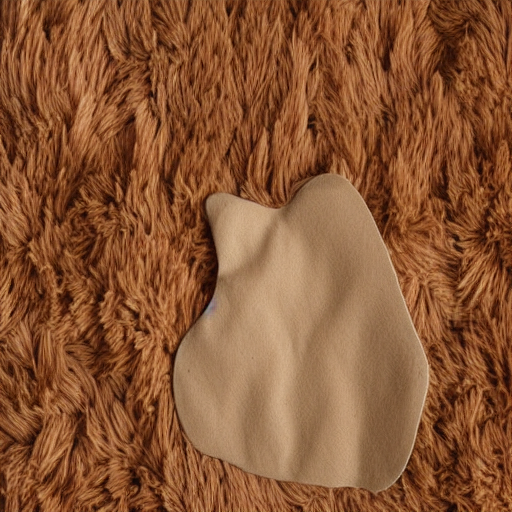} &
        \includegraphics[width=0.185\textwidth]{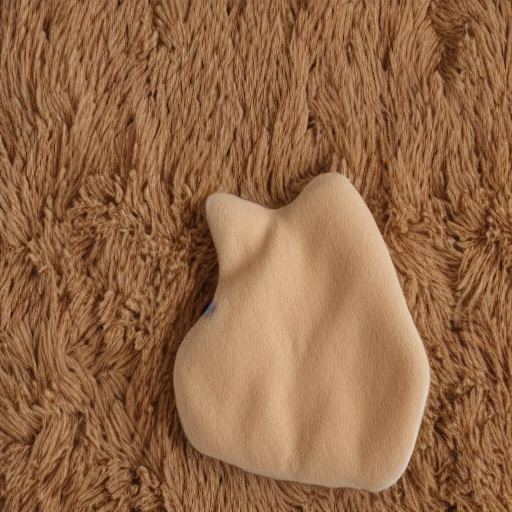} \\
        \includegraphics[width=0.185\textwidth]{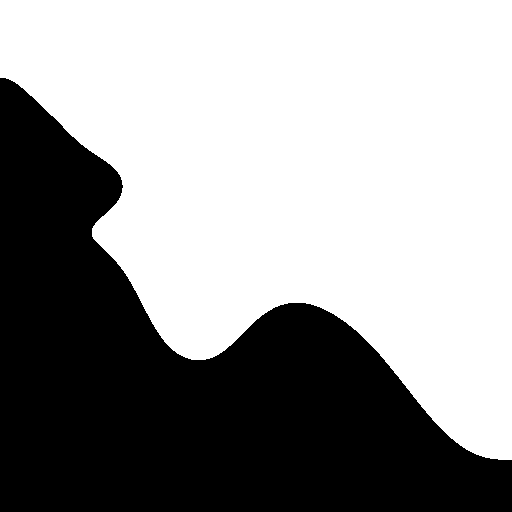} &
        \includegraphics[width=0.185\textwidth]{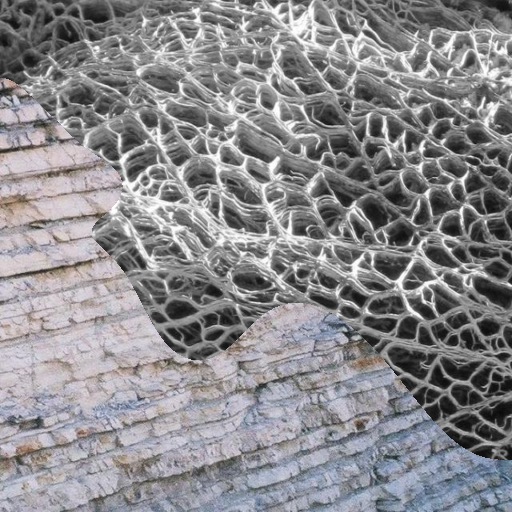} &
        \includegraphics[width=0.185\textwidth]{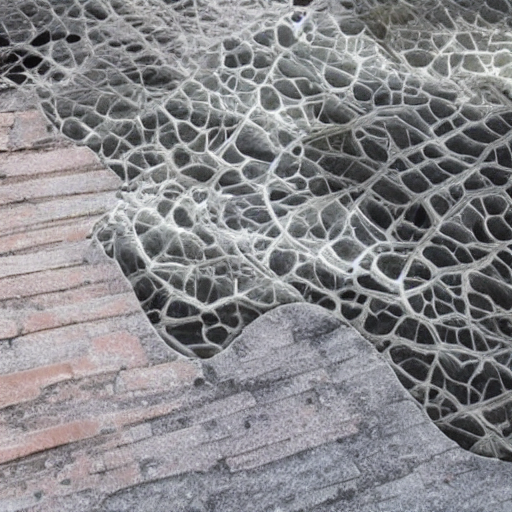} &
        \includegraphics[width=0.185\textwidth]{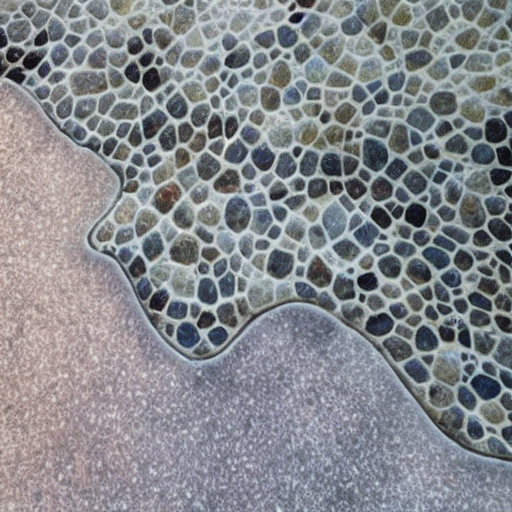} &
        \includegraphics[width=0.185\textwidth]{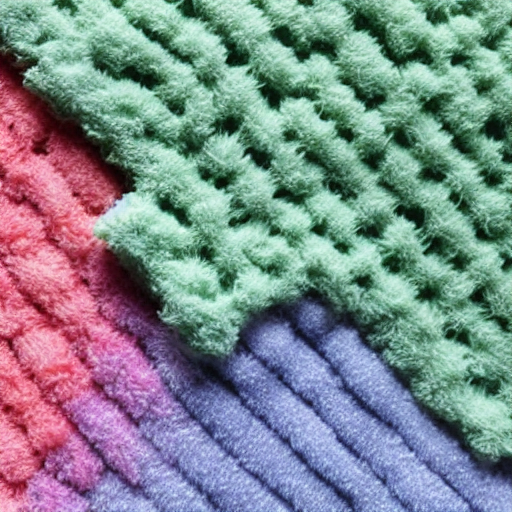}
    \end{tabular}

    \caption{Representative noising strengths \(t'\) for the ControlNet bridge. Late starts preserve source textures but retain artificial seams; early starts smooth transitions but overwrite texture identity. The evaluated bridge set uses the retained generation configuration (\(t'=650\), strength 0.6) rather than all sweep values. ControlNet introduces shadows and transition effects that increase realism while preserving the inherited binary evaluation label.}
    \label{fig:controlnet_t_sweep}
\end{figure*}

The recorded bridge-generation configuration uses an intermediate noising setting, \(t'=650\) with strength 0.6, chosen to preserve texture identity while allowing the renderer to soften the hard seam. Figure~\ref{fig:controlnet_t_sweep} shows a qualitative \(t'\) sweep illustrating this trade-off. Because the route remains synthetic and generator-biased, we treat it only as a synthetic complement, not as a stand-alone texture dataset.

\section{Benchmarks and Evaluation}
\label{sec:benchmarks}

We evaluate four routes, each with a fixed role and evaluator. The point is not to place all routes on one universal scale, but to test whether the same frozen-evidence interpretation holds across natural, synthetic, and bridge regimes.
\paragraph{RWTD.}
RWTD is the historical natural-image anchor and the main fragmented-evidence test. Its weak transitions, repeated distractors, and disconnected same-texture regions make it the route where proposal-bank support and final commitment can diverge most strongly. We use the official label-symmetric evaluator and report mIoU and ARI.
\paragraph{STLD.}
STLD is the historical synthetic anchor. Its exact foregrounds and cleaner geometry test whether the frozen bank already contains a coherent singleton answer rather than requiring extensive assembly. Because the foreground side is canonical, we use direct foreground mIoU and ARI.

\paragraph{ADE20K-selected refined crops.}
The ADE20K-selected route is a compact natural-image complement defined by a fixed crop manifest and paired refined masks, not by TextureSAM predictions. Because the public TextureSAM checkpoint was trained on ADE20K, strict comparison uses the validation slice and matched checkpoint-supported subset. We report partition-invariant mIoU and ARI.

\paragraph{ControlNet bridge.}
The ControlNet bridge is the synthetic complement to STLD. It starts from exact two-region stitched scaffolds with inherited binary labels, then re-renders the boundary with ControlNet-conditioned Stable Diffusion v1.5 to make the seam less trivial than a hard mosaic.

\section{Results}
\label{sec:results}
\begin{table*}[t]
\caption{Matched evaluation across texture routes. SAM2.1-small tests the frozen default readout; TextureSAM is the public-checkpoint rerun; proposal-space recovery is our learned readout above frozen proposals; and feature-space probing is a diagnostic binary readout over frozen features. Metrics are route-matched: RWTD uses the official partition-invariant protocol, ADE20K-selected crops and ControlNet bridge use invariant evaluation, and STLD uses direct foreground evaluation.}
\label{tab:proposal_results}
\centering
\scriptsize
\setlength{\tabcolsep}{4pt}
\renewcommand{\arraystretch}{1.03}
\begin{tabularx}{\textwidth}{@{}Xcc cc cc cc@{}}
\toprule
& \multicolumn{2}{c}{RWTD} 
& \multicolumn{2}{c}{ADE20K-selected crops} 
& \multicolumn{2}{c}{STLD} 
& \multicolumn{2}{c}{ControlNet bridge} \\
\cmidrule(lr){2-3}
\cmidrule(lr){4-5}
\cmidrule(lr){6-7}
\cmidrule(lr){8-9}
Readout or comparator 
& mIoU & ARI 
& mIoU & ARI 
& mIoU & ARI 
& mIoU & ARI \\
\midrule
SAM2.1-small default rerun~\cite{ravi2024sam2}
& 0.16 & 0.22
& 0.17 & 0.10
& 0.37 & 0.53
& 0.28 & 0.22 \\

~\cite{texturesamrepo2026}
& 0.46 & 0.61
& 0.47 & 0.33
& 0.51 & 0.75
& 0.64 & 0.55 \\

Proposal-space recovery
& 0.46 & 0.70
& 0.50 & 0.37
& 0.72 & \textbf{0.77}
& 0.6803 & 0.6039 \\

feature-space probing
& \textbf{0.86} & \textbf{0.75}
& \textbf{0.74} & \textbf{0.55}
& \textbf{0.79} & 0.69
& \textbf{0.84} & \textbf{0.73} \\
\bottomrule
\end{tabularx}
\end{table*}

Table~\ref{tab:proposal_results} tests the paper's central claim: default SAM failure on texture-defined partitions is not texture blindness. Across all four routes, SAM2.1-small is weak, while both frozen evidence spaces expose substantially more usable structure. The feature-space probe asks whether texture organization is present internally; the proposal-space readout asks whether it can be recovered from the external mask bank under a learned commitment rule.

\subsection{Frozen features already organize texture partitions}

The feature-space probe is deliberately diagnostic: it uses a minimal binary readout over frozen features, without proposal reasoning, learned repair, or decoder training. Nevertheless, it reaches strong partition quality on every route: \(0.86/0.75\) on RWTD, \(0.74/0.55\) on ADE20K-selected crops, \(0.79/0.69\) on STLD, and \(0.84/0.73\) on the ControlNet bridge. This rules out the simplest explanation of default failure: the relevant texture signal is not absent from the frozen representation.

This row is not the deployed end-task method. It assumes the binary texture-partition setting directly and bypasses the automatic mask interface. The harder question is whether that information, once externalized as fragmented masks, can be committed into a coherent partition.

\subsection{Frozen proposals contain recoverable but fragmented evidence}

The proposal-space results test that end-task question. On RWTD, proposal-space recovery matches the TextureSAM rerun in overlap while improving partition coherence: mIoU remains about \(0.46\), but ARI increases from \(0.61\) to \(0.70\). This is the expected signature of a consolidation bottleneck: the bank contains useful pieces, but mask selection does not consistently assemble them into the right global partition. STLD shows the complementary regime. With cleaner geometry and a canonical foreground side, the task is less about ambiguous natural grouping and more about selecting a coherent available partition. Proposal-space recovery improves from \(0.51/0.75\) to \(0.72/0.77\), showing that frozen-bank evidence can be sufficient when the target structure aligns with the proposal set.

The broader complements keep the same pattern without carrying the whole claim. On ADE20K-selected crops, proposal-space recovery improves over the TextureSAM rerun from \(0.47/0.33\) to \(0.50/0.37\); on the canonical ControlNet-stitched PTD evaluation archive, from \(0.64/0.55\) to \(0.6803/0.6039\). These routes support the cross-regime interpretation: default masks fail, but frozen evidence often remains recoverable.

Overall, the two paths answer different parts of the same question. The feature probe shows that frozen SAM features encode texture-relevant organization; the proposal route shows that a learned commitment layer can recover competitive end-task partitions from fragmented automatic masks without changing the backbone. The failure mode is therefore better described as readout and commitment mismatch than simple texture blindness.

\section{Discussion and Limitations}
\label{sec:discussion}

\paragraph{Main takeaway.}
The goal of this paper was not to recover texture appearance, but to test whether frozen SAM preserves segmentation-relevant texture evidence. The results support a narrow but important answer: default SAM failure on texture-defined partitions is not sufficient evidence of texture blindness. Texture organization can remain latent in coarse frozen features, and texture-aligned masks or fragments can also appear in the automatic proposal bank. The hardest residual failures are therefore often failures of readout, assembly, and commitment rather than pure representational absence.

\paragraph{Benchmark scope.}
This is why the benchmark design matters. RWTD and STLD remain the historical anchors, while the ADE20K-selected refined-crop companion and the canonical ControlNet-stitched PTD evaluation archive serve as breadth complements. The conclusion comes from agreement across regimes, not from treating any one route as definitive. RWTD stresses fragmented natural evidence, weak transitions, and low-margin commitment; STLD more often reduces to selecting an already coherent proposal; the complements test whether the same interpretation survives under natural-crop and generated-transition settings.

\paragraph{Protocol boundaries.}
The interpretation is protocol-bound. The feature-space probe is diagnostic, not a deployed end-task method. Proposal-space recovery is supervised above a frozen proposal bank: the compatibility model, component scorer, and optional repair policy are trained, although SAM itself and the proposal generator remain frozen. Thus, we do not claim that frozen SAM directly solves texture segmentation, or that the proposal-space readout is unsupervised, training-free, or universal. Comparisons must remain block-local because RWTD, STLD, ADE20K-selected crops, and ControlNet differ in evaluator, subset, coverage, and label-symmetry conventions.

\paragraph{Limitations and open problems.}
Several limitations remain. Generated partition supervision may bias what the consolidation layer finds easy. The feature-space evidence relies on archived probe artifacts rather than a fully matched end-task rerun. The ADE20K-selected complement is curated and compact, and the ControlNet-stitched PTD archive remains synthetic despite being more naturalistic than classical mosaics. Most importantly, RWTD still shows a substantial oracle gap: the frozen bank often contains stronger candidates than the deployed readout recovers. This points to ambiguity-aware commitment, calibrated abstention, and multi-hypothesis outputs above frozen proposal banks as the clearest next steps.

\paragraph{Broader impacts.}
Positive uses include texture-defined segmentation in scientific imaging, materials analysis, remote sensing, and medical or pathology-style workflows where object categories are the wrong abstraction. The frozen-backbone, decision-layer adaptation pattern may also reduce repeated full-model fine-tuning and lower domain-adaptation cost. Risks include ambiguous and domain-dependent boundaries, synthetic-supervision bias, misleading downstream decisions, and possible misuse in sensitive imagery. High-stakes use requires domain validation, uncertainty handling, and human review.
\section{Reproducibility Note}
\label{sec:repro}

Every headline claim is tied to a named artifact, script, or retained summary in the accompanying repository and audit package. RWTD remains evaluation-only throughout the main natural-image route: no main claim depends on RWTD-label training or hyperparameter search. Appendix~\ref{app:artifact_map} makes the retained-artifact boundary explicit for the archived feature-space route and the constructed breadth complements. The manuscript source is self-contained, so compilation does not depend on external figure paths or network-only LaTeX assets.

\section*{Acknowledgments}
This work was supported by the Pazy Foundation.

%\clearpage
{\small
\bibliographystyle{plainnat}
\bibliography{refs}
}

\clearpage
\appendix
\subsection{Remaining gaps concentrate in commitment}

The remaining gaps are informative. The feature-space probe reveals latent structure without settling the final benchmark partition. Proposal-space recovery extracts useful partitions, but the RWTD oracle gap shows that a strong bank still requires better ambiguity handling. The open problem is not total absence of texture evidence; it is commitment under uncertainty: preserving multiple plausible hypotheses, calibrating abstention, and avoiding premature collapse to the wrong texture side.

\begin{table*}[t]
\caption{Compact selector-versus-final contrast on the two historical anchors. RWTD and STLD use different evaluators, so the point is the gap between singleton selection and the full proposal-space readout on each route rather than absolute scale.}
\label{tab:stld_selector_contrast}
\centering
\scriptsize
\begin{tabularx}{\textwidth}{@{}Xcccc@{}}
\toprule
Dataset / subset & Selector mIoU & Selector ARI & Final mIoU & Final ARI \\
\midrule
RWTD common-253 & 0.4512 & 0.5601 & \textbf{0.4645} & \textbf{0.7013} \\
STLD common-182 & \textbf{0.7196} & 0.7731 & 0.7195 & \textbf{0.7791} \\
\bottomrule
\end{tabularx}
\end{table*}

\begin{figure*}[t]
    \centering
    \includegraphics[width=0.98\textwidth]{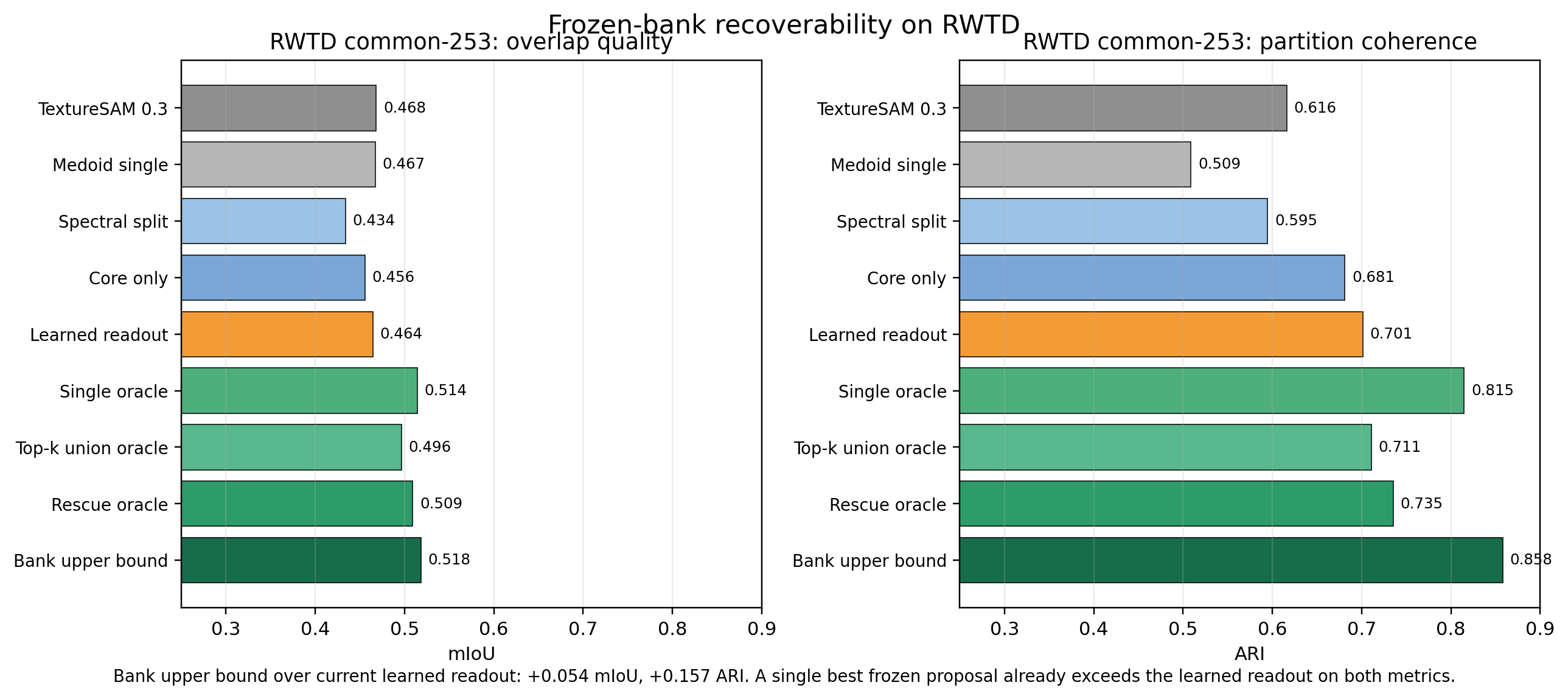}
    \caption{RWTD fragmented-evidence regime. Generic frozen-bank heuristics are weaker than the learned consolidation layer, yet the RWTD bank oracles remain substantially higher still. The answer is not missing evidence, but fragmented evidence plus low-margin commitment.} 
    \label{fig:rwtd_oracle_decomp}
\end{figure*}

\section{Protocol and Claim-Boundary Summary}
\label{app:protocol_summary}

\begin{table*}[t]
\captionsetup{position=top}
\caption{Appendix summary of the two evidence routes. The feature-space probe diagnoses latent texture organization and uses no task-trained decoder. The proposal-bank consolidation readout is learned and supervised in the technical sense, but the learned object is a consolidation rule above a frozen proposal bank rather than a fine-tuned SAM backbone.}
\label{tab:route_protocol}
\centering
\small
\setlength{\tabcolsep}{5pt}
\renewcommand{\arraystretch}{1.16}
\begin{tabularx}{\textwidth}{@{}>{\raggedright\arraybackslash}p{0.17\textwidth}>{\raggedright\arraybackslash}X>{\raggedright\arraybackslash}X>{\raggedright\arraybackslash}p{0.14\textwidth}>{\raggedright\arraybackslash}X@{}}
\toprule
Route & Frozen evidence inspected & Readout & Supervision & Role \\
\midrule
Feature-space probe
& multiscale frozen SAM features
& diagnostic $K=2$ clustering probe
& none
& diagnostic evidence of latent texture organization \\

Proposal-bank consolidation readout (\emph{proposal-space recovery} in tables)
& automatic mask bank
& learned consolidation over proposals
& generated partition supervision
& main end-task evaluative readout \\
\bottomrule
\end{tabularx}
\end{table*}

\begin{table*}[t]
\captionsetup{position=top}
\caption{Appendix claim-boundary summary for interpreting the evaluation. The table records the scope of the supported claim and the conditions under which comparisons are valid; it is not a headline empirical result.}
\label{tab:evaluation_card}
\centering
\footnotesize
\setlength{\tabcolsep}{5pt}
\renewcommand{\arraystretch}{1.12}
\begin{tabularx}{\textwidth}{@{}>{\raggedright\arraybackslash}p{0.22\textwidth}X@{}}
\toprule
Boundary & Statement \\
\midrule
Supported claim
& Frozen SAM preserves enough texture-relevant evidence in frozen multiscale features and generated proposal masks to support texture segmentation under constrained readouts. \\
Unsupported claim
& That frozen SAM directly solves texture segmentation, or that the proposal-bank readout is unsupervised, training-free, or a universal texture segmenter. \\
Evidence spaces
& Frozen multiscale feature space and frozen automatic proposal-bank space. \\
Readout roles
& The feature-space probe is diagnostic. The proposal-bank consolidation readout (\emph{proposal-space recovery} in tables) is the matched end-task evaluator. \\
Supervision source
& Generated partition supervision for RWTD, ADE20K-selected texture crops, and the canonical ControlNet-stitched PTD evaluation archive; Brodatz-native synthetic supervision for STLD; no SAM fine-tuning, no proposal-generator retraining, and no target-domain dense-mask training on the main natural-image route. \\
Valid comparison scope
& Comparisons are block-local: fixed dataset, subset, evaluator, coverage, and label-symmetry conventions. \\
Main unresolved failure mode
& Ambiguity-aware commitment above strong but fragmented proposal banks, especially on RWTD. \\
\bottomrule
\end{tabularx}
\end{table*}

\begin{table*}[t]
\captionsetup{position=top}
\caption{Compact reproducibility summary for the four central evaluation routes.}
\label{tab:compact_repro}
\centering
\small
\begin{tabularx}{\textwidth}{@{}p{0.30\textwidth}X@{}}
\toprule
Item & Value \\
\midrule
External training data & PTD only for RWTD, the ADE20K-selected complement, and the ControlNet bridge complement; Brodatz-native synthetic supervision for STLD \\
Encoder training data & 80{,}000 train / 20{,}000 val images per synthetic source \\
RWTD benchmark usage & Evaluation only; no RWTD-label training or hyperparameter search \\
Proposal source & Frozen SAM-style automatic masks \\
RWTD / STLD deployed model & Masked-region encoder with ConvNeXt-Tiny PTD backbone and learned merge/core/repair modules. Legacy \texttt{swinb} appears only in artifact/folder labels and does not define the model claim. \\
ADE20K / ControlNet deployed model & PTD-trained \texttt{small\_pre\_ring\_hi} Stage-A bundle \\
Primary evaluator & Official upstream no-aggregation evaluator on RWTD; route-matched evaluators elsewhere \\
Comparator family & Raw SAM2.1-small plus reproduced/public TextureSAM checkpoints \\
\bottomrule
\end{tabularx}
\end{table*}

\section{Implementation Details and Practicality}
\label{app:implementation_details}
\suppressfloats[t]

\begin{table}[t]
\caption{Compact pseudocode for the deployed proposal-space consolidation layer. All non-RWTD routes stop after Step~4 because they do not use the RWTD rescue layer.}
\label{tab:algorithm_box}
\centering
\footnotesize
\setlength{\tabcolsep}{0pt}
\renewcommand{\arraystretch}{1.02}
\begin{tabularx}{\columnwidth}{@{}X@{}}
\toprule
\textbf{Algorithm 1: proposal-space inference over the frozen mask bank for image $x$} \\
\midrule
1. Load the frozen prompt-bank proposals $\mathcal{P}(x)$ and compute proposal descriptors $d_i$ for each $m_i \in \mathcal{P}(x)$. \\
2. Score adjacent proposal pairs with $s(i,j)$ and threshold the compatibility graph. \\
3. Form candidate merged regions from the graph connected components. \\
4. Score each candidate with $q(C)$ and choose the conservative core $C^\star = \arg\max_C q(C)$. \\
5. If the route is RWTD, build dense rescue candidates $R_k$ from the auxiliary bank $\mathcal{P}^{+}(x)$. \\
6. Score each $(C^\star, R_k)$ with the safe-switch classifier $p_{\mathrm{safe}}(k)$ and gain regressor $g_{\mathrm{gain}}(k)$. \\
7. Output the positive rescue candidate with highest predicted gain, if any; otherwise output $C^\star$. \\
\bottomrule
\end{tabularx}
\end{table}

\begin{table*}[t]
\caption{Descriptor ablation for the descriptor-sensitive Stage-A commitment layer, with the proposal bank, evaluator, and model family held fixed. RWTD is reported at Stage-A because the dense rescue layer operates on candidate masks and is unchanged across descriptor variants. Hybrid descriptors are clearly strongest on RWTD, while STLD is less sensitive to the descriptor choice and even slightly favors handcrafted-only on ARI.}
\label{tab:descriptor_ablation}
\centering
\scriptsize
\setlength{\tabcolsep}{4pt}
\renewcommand{\arraystretch}{0.98}
\resizebox{\textwidth}{!}{%
\begin{tabular}{@{}llcccccc@{}}
\toprule
Benchmark / subset & Output & \multicolumn{2}{c}{Handcrafted-only} & \multicolumn{2}{c}{PTD-only} & \multicolumn{2}{c}{Hybrid} \\
 &  & mIoU & ARI & mIoU & ARI & mIoU & ARI \\
\midrule
RWTD common-253 & Stage-A official & 0.4469 & 0.6556 & 0.4486 & 0.6505 & 0.4558 & 0.6812 \\
RWTD full-256 & Stage-A official & 0.4454 & 0.6533 & 0.4472 & 0.6484 & 0.4543 & 0.6786 \\
STLD common-182 & Stage-A direct & 0.7045 & 0.7808 & 0.6698 & 0.7740 & 0.7195 & 0.7791 \\
STLD all-200 & Stage-A direct & 0.6570 & 0.7265 & 0.6253 & 0.7203 & 0.6705 & 0.7249 \\
\bottomrule
\end{tabular}
}
\end{table*}

\begin{table*}[t]
\caption{Practicality profile of the deployed proposal-space consolidation layer. RWTD timings come from measured full-dataset GPU profiles of the same frozen proposal bank pipeline. STLD shares the same Stage-A architecture and bank statistics, but no separate runtime profile was recorded in the repository.}
\label{tab:practicality}
\centering
\scriptsize
\setlength{\tabcolsep}{3pt}
\renewcommand{\arraystretch}{0.98}
\begin{tabularx}{\textwidth}{@{}p{0.14\textwidth}ccccccX@{}}
\toprule
Route & Prompt props & Merged comps & Dense props & Rescue cand. & s / img & Peak RSS & Reading \\
\midrule
RWTD Stage-A & 3.84 & 2.64 & -- & -- & 0.025 & 430 MiB & prompt-bank commitment only \\
RWTD full & 3.84 & 2.64 & 38.05 & 11.72 & 0.378 & 637 MiB & dense rescue evaluated on the full set; 21/256 images switch \\
STLD Stage-A & 1.78 & 2.32 & -- & -- & -- & -- & no rescue path; no separate runtime profile recorded \\
\bottomrule
\end{tabularx}
\end{table*}

\subsection{RWTD and STLD teach different failure modes}

\Cref{tab:stld_selector_contrast} makes the contrast compactly visible: RWTD gains substantially from the full commitment stack over the learned singleton selector, whereas STLD is already near-parity and differs mainly in coherence.

RWTD is the benchmark where the proposal-space story is most informative. The oracle analysis shows that the frozen bank often already contains strong answers, but those answers are distributed across fragments, low-margin alternatives, or under-covered cores. Learned top-1 selection helps, but it still trails the full readout by a wide ARI margin. RWTD is therefore a fragmented-evidence and commitment-limited regime, not simply a missing-evidence regime.

STLD behaves differently. There, a learned singleton selector already nearly matches the full proposal-space readout, which means the frozen bank often contains a coherent answer that mostly needs to be chosen rather than assembled. This contrast is one of the main scientific outputs of the analysis. Frozen features may already organize texture structure internally, but what the proposal bank externalizes depends on the regime: natural fragmented scenes stress commitment, while cleaner synthetic scenes often reduce to singleton selection.

\begin{table}[t]
\centering
\small
\begin{tabular}{lcccc}
\toprule
Route & Probe variant & mIoU & ARI & NMI \\
\midrule
DeTexture ADE20K & feature clustering, oracle-$K$ & 0.7331 & 0.6749 & 0.6757 \\
\bottomrule
\end{tabular}
\caption{K-known feature-space diagnostic on DeTexture ADE20K. The ground truth supplies only the number of regions \(K\); clustering is still fit on frozen features without using ground-truth geometry or pixel labels. This result is not merged into the binary Table~1 protocol.}
\label{tab:detexture_oracle_k_feature_probe}
\end{table}

\section{Feature-Space Evidence}
\label{app:feature_provenance}

The feature-space route is reported under its archived probe protocol and read as evidence for where texture structure lives in frozen features, not as a state-of-the-art foundation model.

\begin{figure*}[t]
    \centering
    \includegraphics[width=0.98\textwidth]{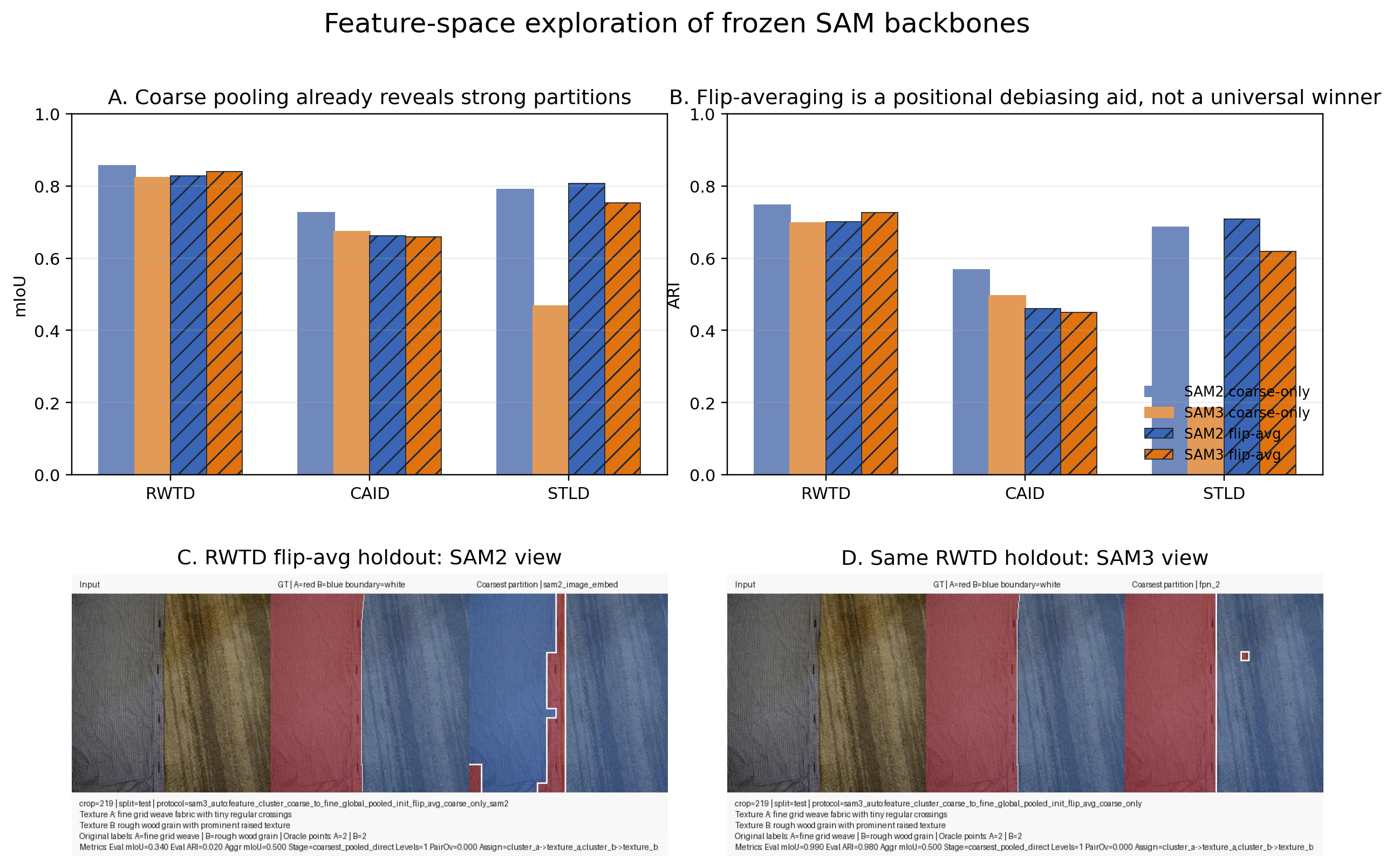}
    \caption{Feature-space route summary under the archived probe protocol. Top: coarse-only and flip-averaged summaries across RWTD and STLD. Bottom: representative RWTD holdouts showing that coarse clustering can recover plausible binary partitions while remaining sensitive to positional bias. The figure is read as evidence for what frozen features reveal, not as a matched leaderboard comparison.}
    \label{fig:feature_summary_app}
\end{figure*}

\begin{figure*}[t]
    \centering
    \includegraphics[width=0.98\textwidth]{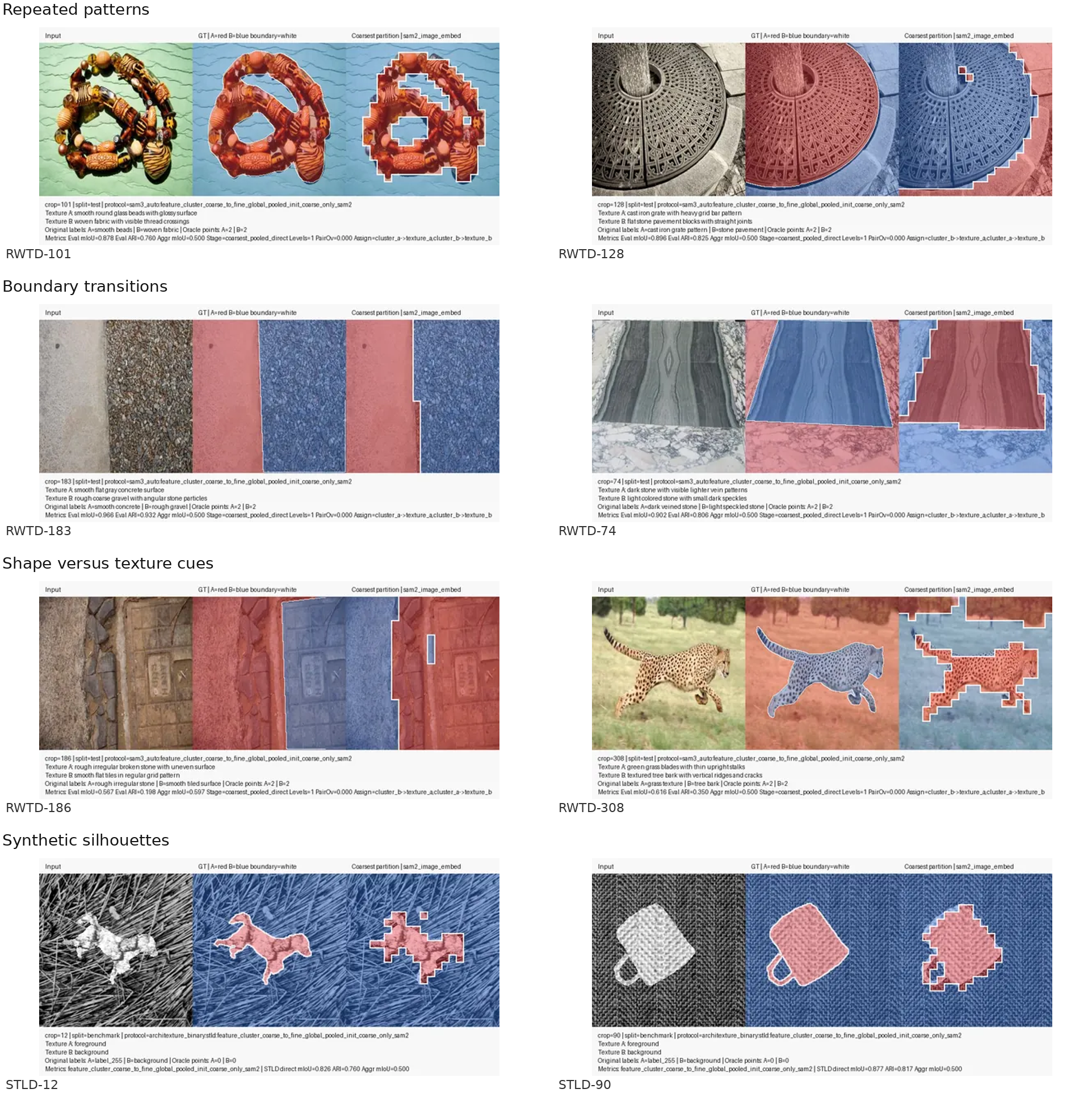}
    \caption{Feature-space route qualitative gallery. The examples span repeated-pattern scenes, gradual texture transitions, shape-versus-texture cases, and synthetic silhouettes. The annulus example in RWTD-101 is representative of the strongest qualitative evidence: even a frozen coarse probe can recover a topologically nontrivial partition without retraining the backbone.}
    \label{fig:feature_gallery}
\end{figure*}

\begin{figure*}[t]
    \centering
    \includegraphics[width=0.98\textwidth]{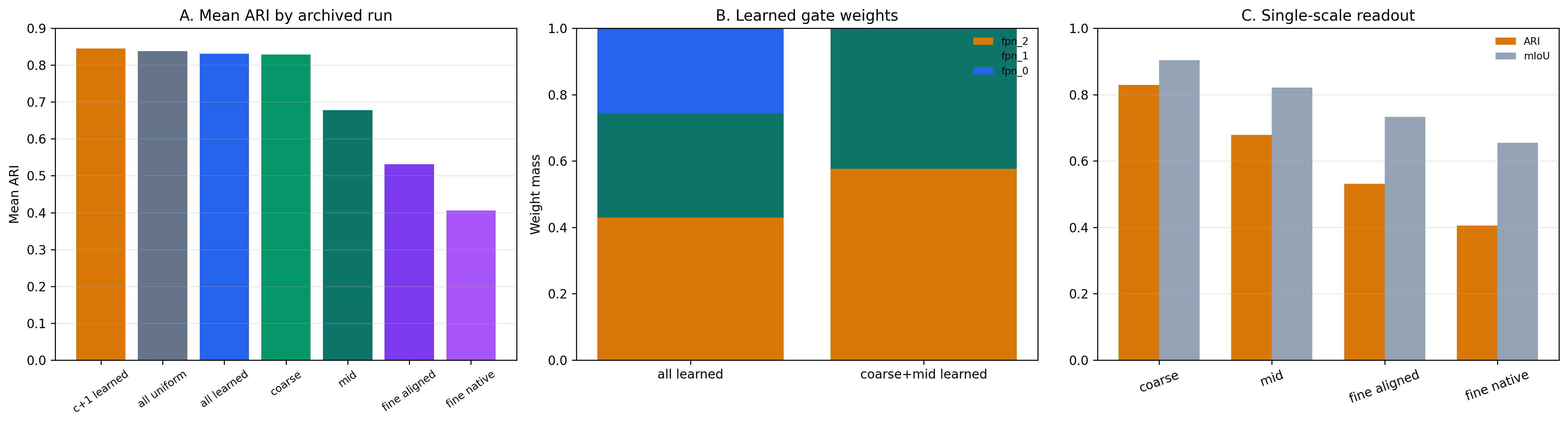}
    \caption{Feature-space route scale and positionality diagnostics. Left: mean ARI by run. Middle: learned all-scale and coarse-plus-mid gate weights still place most mass on the coarsest level. Right: single-scale readouts show that coarse features dominate, while native fine-grid clustering is the weakest control. These runs are read as archived probe evidence rather than as matched benchmark comparisons.}
    \label{fig:feature_scale_diag}
\end{figure*}

\paragraph{Interpretation.} The appendix diagnostics point in one direction. Coarse frozen features already recover much of the binary partition on the repository-backed probe, learned all-scale fusion still prefers the coarsest level, and the finest native grid is the least reliable setting. This is enough for the route's role: frozen SAM features can contain texture-aware structure, even though the feature-space probe is not the matched benchmark route used for RWTD and STLD headline comparisons.

\begin{table}[t]
\caption{Archived frozen-feature parity summary from the site bundle. Each cell is reported as mIoU / ARI. This table is reported under the archived feature-probe protocol and is not merged into the matched proposal-space benchmark tables.}
\label{tab:feature_parity}
\centering
\scriptsize
\setlength{\tabcolsep}{4pt}
\renewcommand{\arraystretch}{0.98}
\resizebox{\columnwidth}{!}{%
\begin{tabular}{@{}llcccc@{}}
\toprule
Dataset & Eval view & SAM2 coarse-only & SAM3 coarse-only & SAM2 flip-avg & SAM3 flip-avg \\
\midrule
RWTD & partition-invariant & 0.8561 / 0.7472 & 0.8235 / 0.6984 & 0.8278 / 0.7013 & 0.8397 / 0.7258 \\
CAID & partition-invariant & 0.7263 / 0.5673 & 0.6743 / 0.4956 & 0.6620 / 0.4605 & 0.6588 / 0.4507 \\
STLD & direct foreground & 0.7908 / 0.6864 & 0.4682 / 0.1859 & 0.8064 / 0.7089 & 0.7534 / 0.6194 \\
\bottomrule
\end{tabular}
}
\end{table}

\subsection{DeTexture multi-label oracle-$K$ diagnostic}

The multi-label DeTexture companion is a useful stress test because it is not a binary split. Here the ground truth supplies only the region count \(K\), and the clustering readout must separate multiple texture regions in frozen feature space without pixel supervision. The point is not to claim a new benchmark result; it is to show that the same frozen feature-route can represent multi-label texture structure when the number of regions is known.

\begin{figure*}[t]
    \centering
    \begin{tabular}{ccc}
    \includegraphics[width=0.8\textwidth]{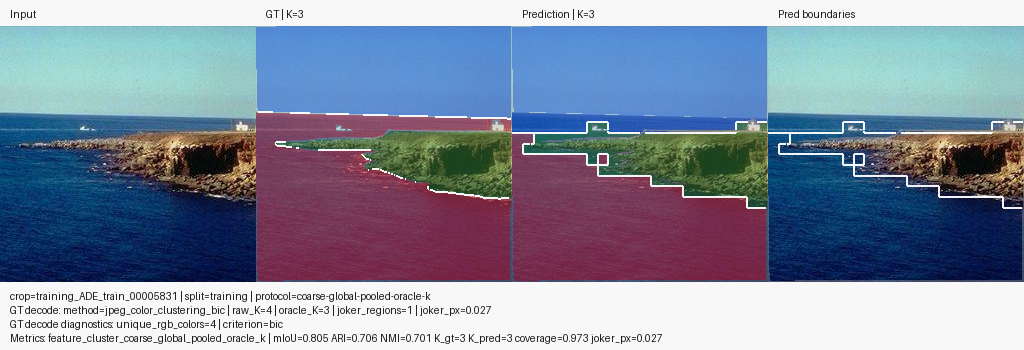} \\
        \includegraphics[width=0.8\textwidth]{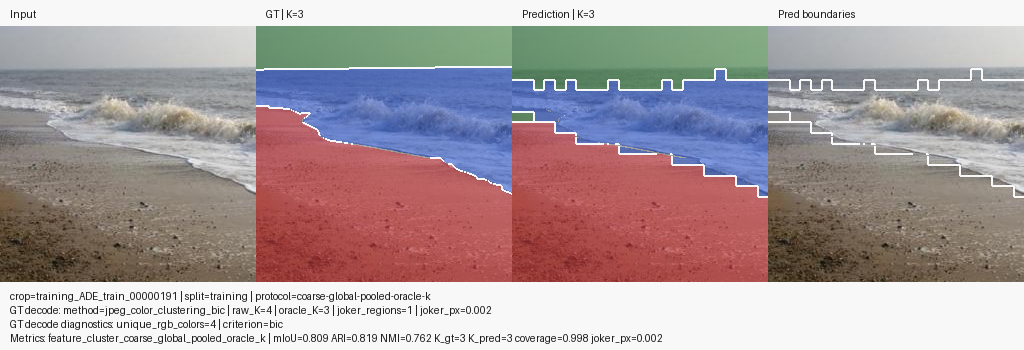}\\
        \includegraphics[width=0.8\textwidth]{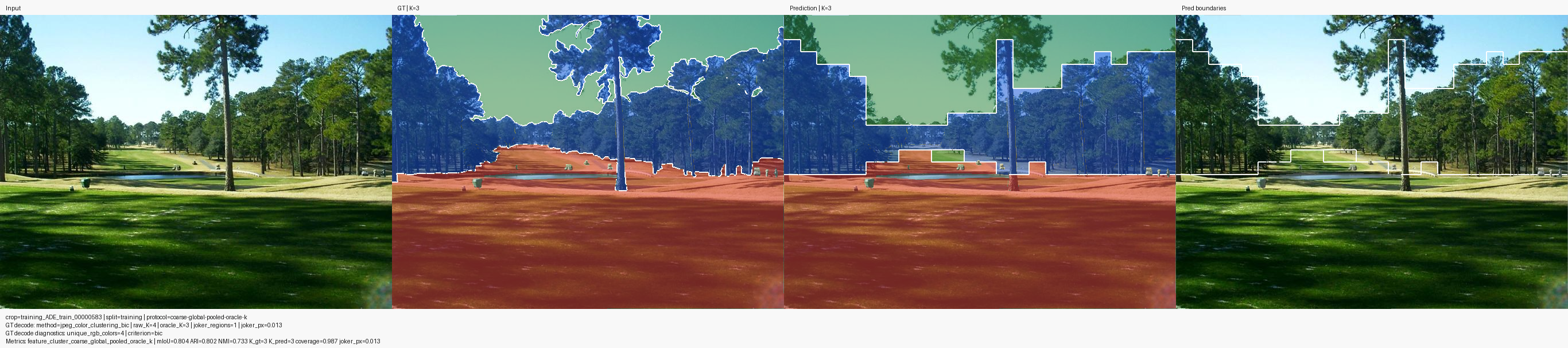} \\
        \includegraphics[width=0.8\textwidth]{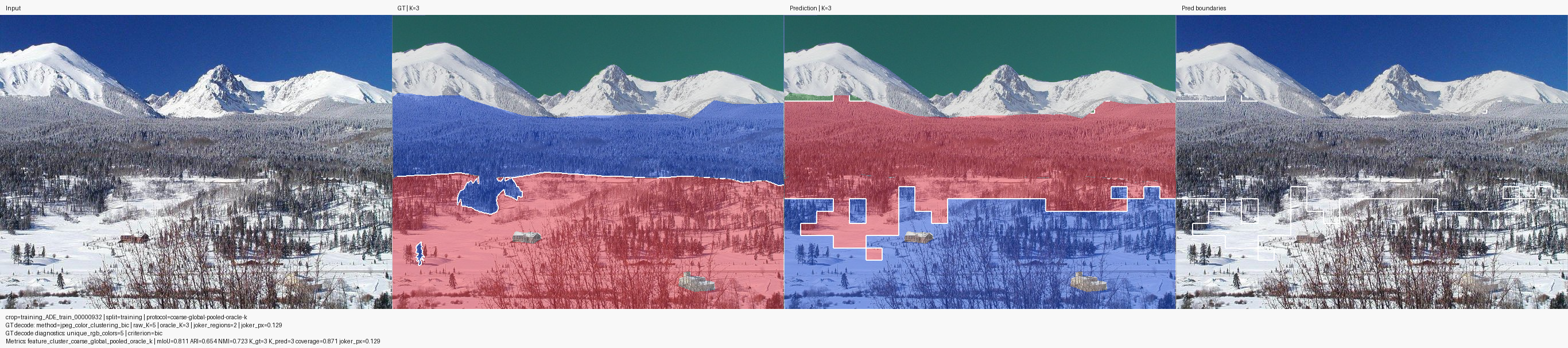} \\
        \includegraphics[width=0.8\textwidth]{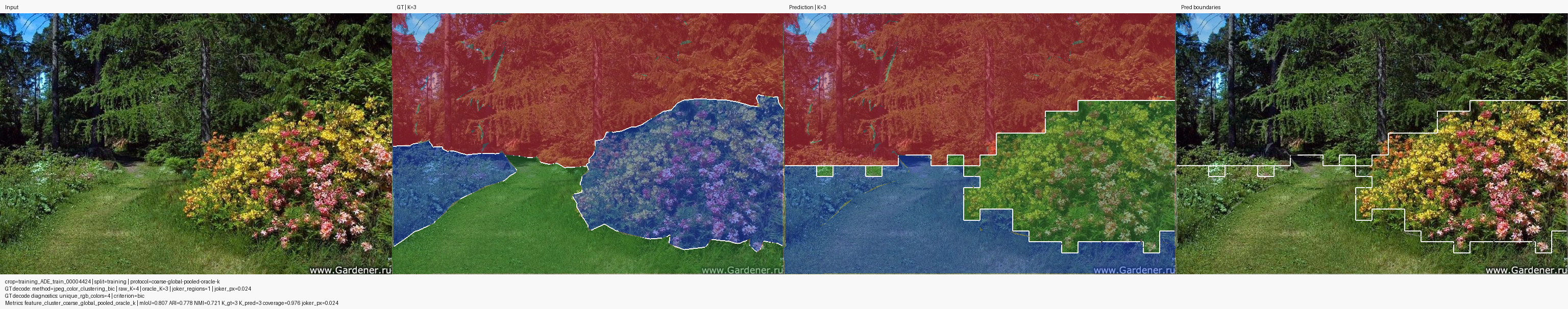} \\
        \includegraphics[width=0.8\textwidth]{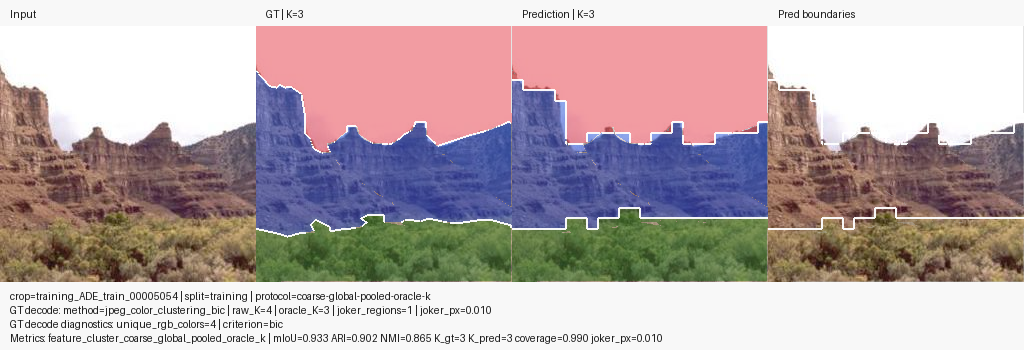} \\
        
    \end{tabular}
    \caption{DeTexture multi-label feature-space probe under oracle-$K$. The retained companion crops show that the same frozen-feature clustering readout can separate multiple texture regions, not only binary partitions, when the region count is known. The montage is qualitative by design: it illustrates multi-label texture organization in frozen features without using pixel labels or ground-truth geometry.}
    \label{fig:detexture_multilabel_feature}
\end{figure*}

\begin{table}[t]
\caption{DeTexture multi-label feature-space diagnostic under oracle-$K$. The ground truth supplies only the number of regions \(K\); clustering is still fit on frozen features without using ground-truth geometry or pixel labels. This appendix view shows that the feature-space route can handle multi-label structure, not just binary partitions.}
\label{tab:detexture_multilabel_feature}
\centering
\small
\begin{tabular}{lcccc}
\toprule
Route & Probe variant & mIoU & ARI & NMI \\
\midrule
DeTexture multi-label & feature clustering, oracle-$K$ & 0.7331 & 0.6749 & 0.6757 \\
\bottomrule
\end{tabular}
\end{table}

\section{RWTD Oracle and Failure Details}
\label{app:rwtd_diag}

\begin{figure*}[t]
    \centering
    \includegraphics[width=\textwidth]{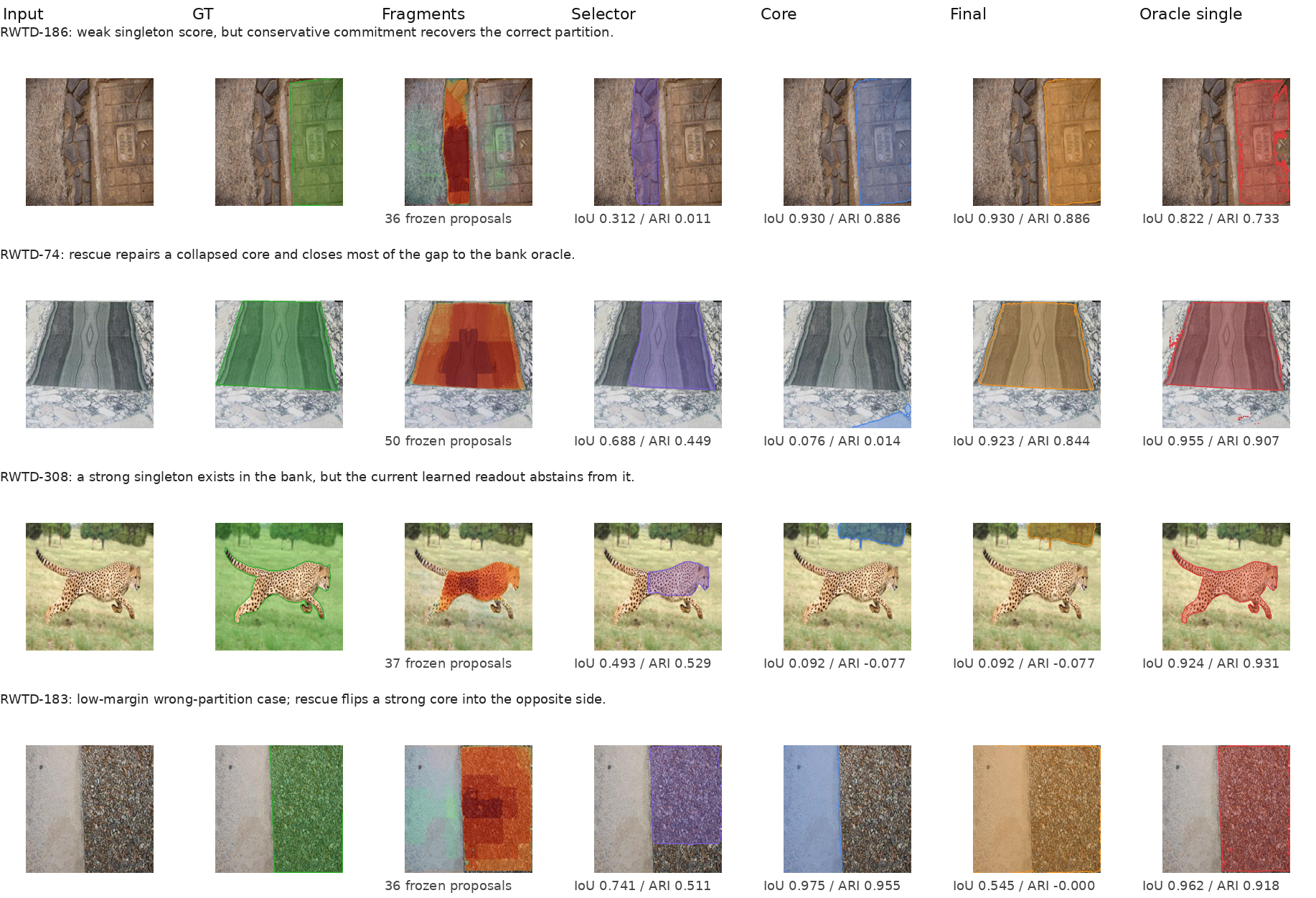}
    \caption{Proposal-space recovery case studies on RWTD. The panels compare selector, core, final, and oracle views so that the route's remaining difficulty is visible image by image. The bank often contains a strong answer, but the deployed readout still needs the right commitment decision to reach it. Rescue helps on some under-coverage cases, while low-margin wrong-partition failures remain brittle.}
    \label{fig:rwtd_case_gallery}
\end{figure*}

RWTD is where the preserved-evidence claim is most demanding. The proposal bank is strong, but the current consolidation layer still leaves a large oracle gap. \Cref{fig:rwtd_case_gallery} shows four recurring patterns. First, top-1 proposal scoring can miss cases where conservative component commitment succeeds. Second, rescue can repair an under-covered core. Third, a strong singleton can already exist in the bank even when the learned selector abstains from it. Fourth, the remaining wrong-partition failures are low-margin commitment mistakes rather than proposal-bank misses.

\begin{figure*}[t]
    \centering
    \includegraphics[width=0.92\textwidth]{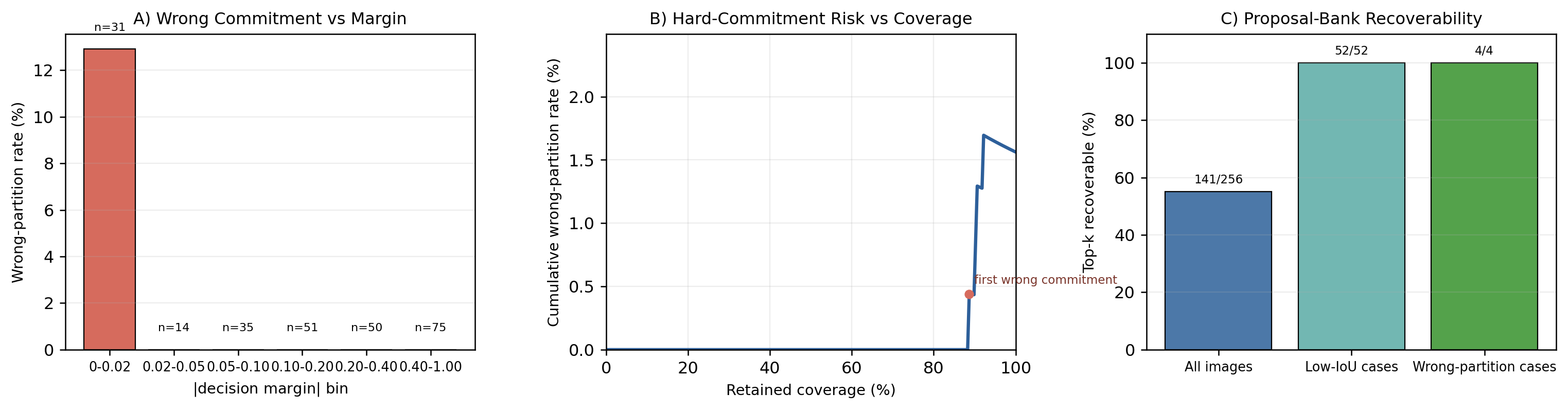}
    \caption{Proposal-space ambiguity diagnostics on RWTD. Wrong-partition cases concentrate in the lowest-margin regime, and top-$k$ bank coverage remains high where the final commitment is most brittle. The unresolved issue is not absence of texture evidence in the generated masks, but ambiguity-aware commitment above that evidence.}
    \label{fig:ambiguity_commitment}
\end{figure*}

\paragraph{How to read the oracle gap.} The bank upper bound is not a deployable readout: it assumes access to the ground-truth partition when choosing among proposals, merged components, and rescue candidates. Its role is diagnostic. On RWTD, the gap between the deployed readout and that bound shows that the frozen bank already contains more usable texture evidence than the current consolidation layer recovers. The learned single-selector baseline narrows only part of that gap, which is why RWTD still needs structured commitment above fragmented proposals.

\begin{table}[t]
\caption{RWTD oracle and commitment breakdown on the matched common-253 subset. The learned top-1 selector is stronger than the medoid single baseline, but it still trails the full commitment stack by a wide ARI margin. The bank oracles remain far higher still. The omitted compatible-component oracle performs poorly (0.4748 mIoU / 0.2497 ARI), which is evidence that compatibility merging alone is not sufficient.}
\label{tab:oracle_recoverability}
\centering
\scriptsize
\begin{tabular}{@{}lcc@{}}
\toprule
RWTD common-253 readout / oracle & mIoU & ARI \\
\midrule
Medoid single & 0.4673 & 0.5087 \\
Learned single selector & 0.4512 & 0.5601 \\
Spectral split & 0.4339 & 0.5948 \\
Agglomerative split & 0.4354 & 0.5898 \\
Greedy merge & 0.4442 & 0.5197 \\
Core only & 0.4558 & 0.6812 \\
Proposal-space recovery final & 0.4645 & 0.7013 \\
Single frozen-proposal oracle & 0.5142 & 0.8146 \\
Top-$k$ union oracle & 0.4964 & 0.7107 \\
Rescue candidate oracle & 0.5093 & 0.7352 \\
Bank upper bound & 0.5183 & 0.8580 \\
\bottomrule
\end{tabular}
\end{table}

\begin{table*}[t]
\caption{RWTD full-256 failure-audit summary for the deployed consolidation rule. The remaining weakness is ambiguity-aware commitment rather than proposal-bank absence.}
\label{tab:rwtd_audit}
\centering
\small
\begin{tabularx}{\textwidth}{@{}Xccccccc@{}}
\toprule
Setting & Resc. & Off-succ. & Hurt-any & Wrong & Unsafe & Mean $\Delta$IoU & Top-$k$ recov. \\
\midrule
Baseline decision rule & 21 & 61.9\% & 38.1\% & 4 & 6 & +0.2247 & 141/256 \\
\bottomrule
\end{tabularx}
\end{table*}

\begin{table*}[t]
\caption{Frozen proposal-source transfer stress test. Consolidation weights and thresholds remain fixed; only the rescue proposal bank changes. This is evidence for modularity of the consolidation layer, not a standalone benchmark claim.}
\label{tab:proposal_swap}
\centering
\scriptsize
\begin{tabularx}{\textwidth}{@{}Xcc@{}}
\toprule
Dense rescue proposal source (frozen) & Full-256 mIoU / ARI & Common-253 mIoU / ARI \\
\midrule
Released dense bank & 0.4611 / 0.6966 & 0.4645 / 0.7013 \\
SAM2.1-large source swap & 0.5003 / 0.7048 & 0.5005 / 0.7106 \\
\bottomrule
\end{tabularx}
\end{table*}

\begin{table*}[t]
\caption{Paired interval support for the main proposal-space comparisons. RWTD uses the official overlap-instance view because the published no-aggregation metric averages GT instances overlapped by each compared system rather than simple per-image means. STLD intervals are paired per-image bootstraps under the direct-foreground evaluator.}
\label{tab:bootstrap_support}
\centering
\scriptsize
\setlength{\tabcolsep}{5pt}
\renewcommand{\arraystretch}{0.98}
\begin{tabularx}{\textwidth}{@{}p{0.22\textwidth}p{0.17\textwidth}cc@{}}
\toprule
Comparison & Paired units & $\Delta$ mIoU (95\% interval) & $\Delta$ ARI (95\% interval) \\
\midrule
RWTD common-253, proposal-space recovery $-$ TextureSAM & 426 overlapped GT instances & $+0.0405$ [$0.0031$, $0.0764$] & $+0.0995$ [$0.0721$, $0.1270$] \\
STLD all-200, proposal-space recovery $-$ TextureSAM & 200 images & $+0.2028$ [$0.1555$, $0.2486$] & $+0.0400$ [$0.0049$, $0.0754$] \\
STLD common-182, proposal-space recovery $-$ TextureSAM & 182 images & $+0.2055$ [$0.1558$, $0.2518$] & $+0.0265$ [$-0.0097$, $0.0605$] \\
\bottomrule
\end{tabularx}
\end{table*}

\section{STLD Supporting Oracle Analysis}
\label{app:stld_support}

STLD plays a different supporting role. The oracle analysis still shows extra headroom in the bank, but the learned top-1 selector is already close to the full proposal-space readout on the common-182 subset. In practice, many STLD images contain a coherent singleton answer, so the main remaining gain is modest coherence cleanup rather than large-scale fragment commitment.

\begin{figure*}[t]
    \centering
    \includegraphics[width=0.96\textwidth]{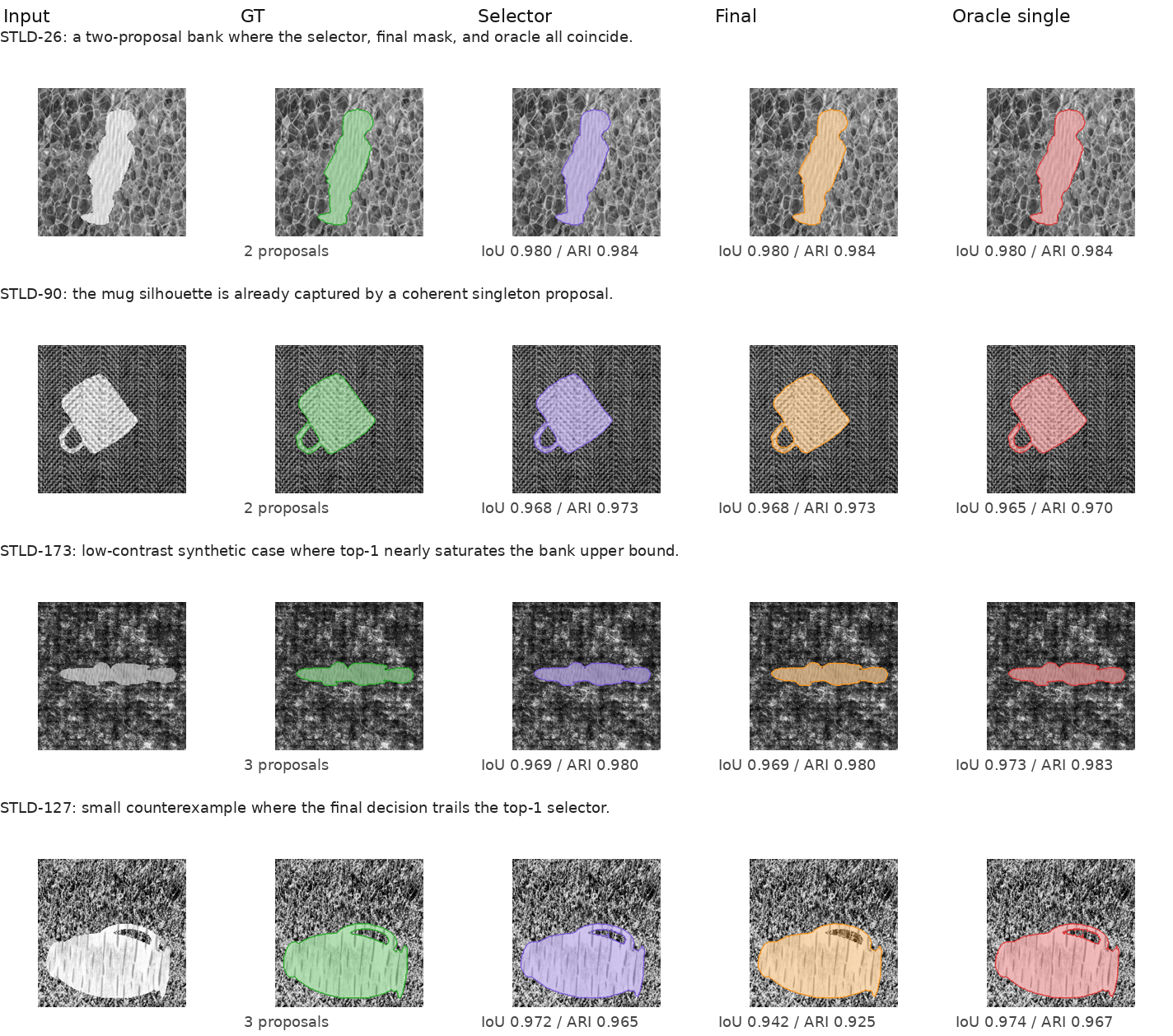}
    \caption{Proposal-space recovery case studies on STLD. Most examples already have a coherent singleton proposal, so the learned selector, final deployment, and single-proposal oracle are nearly indistinguishable. The last row shows a useful counterexample where the selector slightly outruns the final decision. Compared with RWTD, STLD more often presents a singleton-selection problem than a fragmented-commitment problem.}
    \label{fig:stld_case_gallery}
\end{figure*}

\begin{table*}[t]
\caption{Supporting STLD oracle analysis under the direct-foreground evaluator. The learned top-1 selector already recovers much of the overlap on STLD, but the full proposal-space readout remains slightly stronger on coherence and still trails the bank oracles.}
\label{tab:stld_oracles}
\centering
\scriptsize
\setlength{\tabcolsep}{5pt}
\renewcommand{\arraystretch}{0.98}
\begin{tabular}{@{}lcccc@{}}
\toprule
STLD readout / oracle & All-200 mIoU & All-200 ARI & Common-182 mIoU & Common-182 ARI \\
\midrule
Proposal-space recovery final & 0.6705 & 0.7249 & 0.7195 & 0.7791 \\
Learned single selector & 0.6548 & 0.7036 & 0.7196 & 0.7731 \\
Single frozen-proposal oracle & 0.7375 & 0.7684 & 0.8105 & 0.8444 \\
Compatible-component oracle & 0.3950 & 0.3598 & 0.4341 & 0.3953 \\
Top-$k$ union oracle & 0.1257 & 0.0289 & 0.1381 & 0.0317 \\
Bank upper bound & 0.7574 & 0.7962 & 0.8149 & 0.8574 \\
\bottomrule
\end{tabular}
\end{table*}

\section{Complementary Benchmark Details}
\label{app:benchmark_complements}

\subsection{ADE20K-Selected Natural Complement}
\label{app:detexture_secondary}

56 paired crop tests from refined ADE20K crops with fixed manifest and masks. Validation-slice comparison avoids training leakage.

\begin{figure*}[t]
    \centering
    \includegraphics[width=0.98\textwidth]{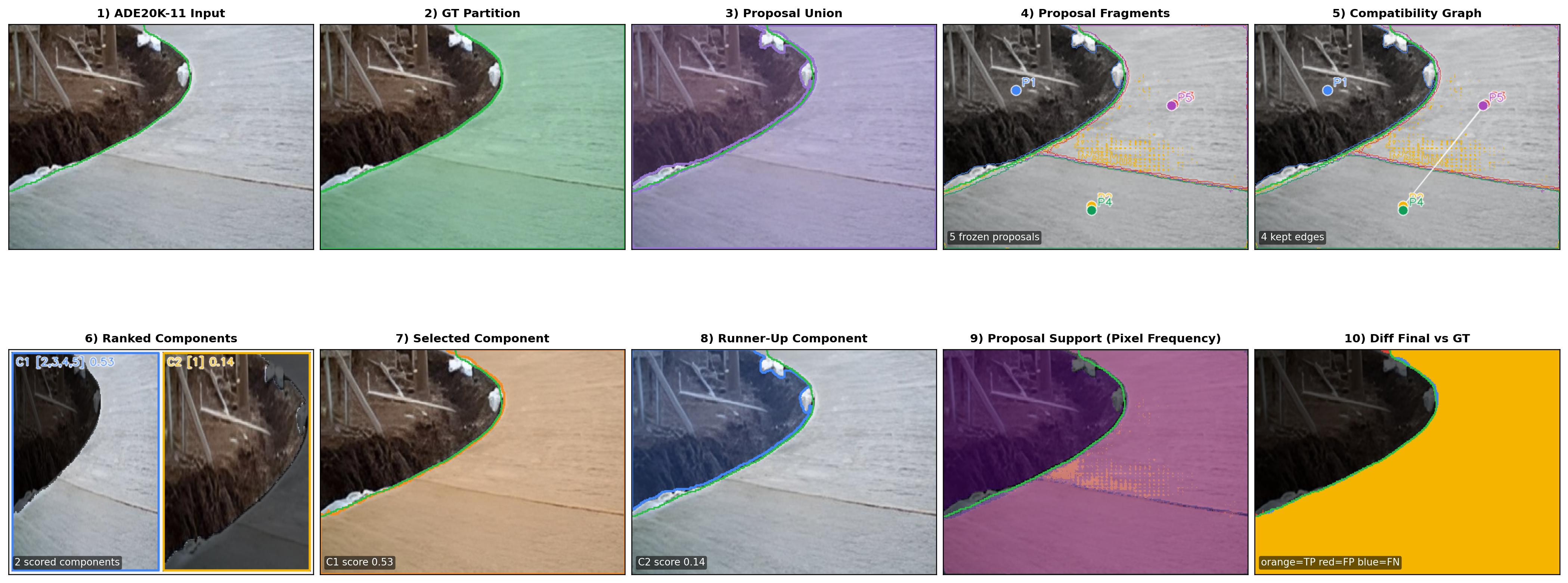}
    \caption{ADE20K-selected complement: proposal-space consolidation example.}
    \label{fig:detexture_app}
\end{figure*}

\begin{table*}[t]
\caption{ADE20K-selected natural complement. The route is defined by a companion refined-crop artifact included with the supplementary package: a fixed crop manifest and paired refined masks over ADE20K images. In total it contains 56 paired crop tests, or 56 binary evaluation images. The crop and mask definitions are independent of TextureSAM predictions; TextureSAM appears here only as a reproduced comparator. To avoid ADE20K training leakage in that comparison, the strict matched view uses the validation slice. The main text emphasizes the invariant columns because the two texture sides are label-symmetric, while this appendix table preserves the direct view and the full-versus-matched coverage context.}
\label{tab:detexture_secondary}
\centering
\scriptsize
\begin{tabularx}{\textwidth}{@{}Xlcc@{}}
\toprule
Readout / comparator & View / coverage & Direct mIoU / ARI & Invariant mIoU / ARI \\
\midrule
SAM2.1-small original rerun~\cite{ravi2024sam2} & validation full set, 56/56 & 0.1263 / 0.1379 & 0.1665 / 0.1015 \\
TextureSAM public checkpoint rerun~\cite{texturesamrepo2026} & validation full set, 54/56 & 0.2899 / 0.3649 & 0.4547 / 0.3187 \\
Proposal-space recovery (Stage-A) & validation full set, 56/56 & 0.2544 / 0.3675 & 0.5008 / 0.3675 \\
TextureSAM public checkpoint rerun~\cite{texturesamrepo2026} & validation matched subset, 54/54 & 0.3007 / 0.3785 & 0.4715 / 0.3306 \\
Proposal-space recovery (Stage-A) & validation matched subset, 54/54 & 0.2597 / 0.3762 & 0.5043 / 0.3762 \\
\bottomrule
\end{tabularx}
\end{table*}

\subsection{ControlNet Bridge Synthetic Complement}
\label{app:bridge_secondary}

1,742 retained evaluation images; exact two-region synthetic partitions with label-symmetric partition-invariant metrics.

\begin{figure*}[t]
    \centering
    \begin{tabular}{ccc}
        \includegraphics[width=0.55\textwidth]{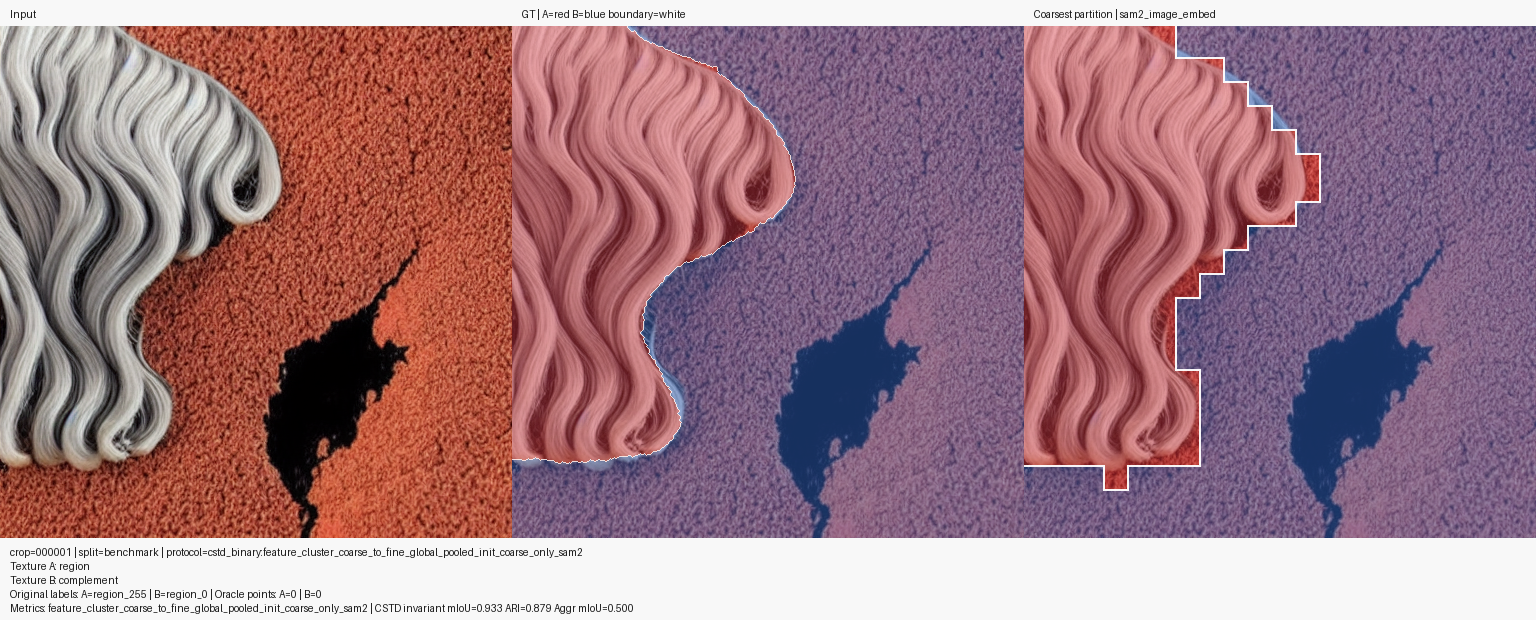} \\
        \includegraphics[width=0.55\textwidth]{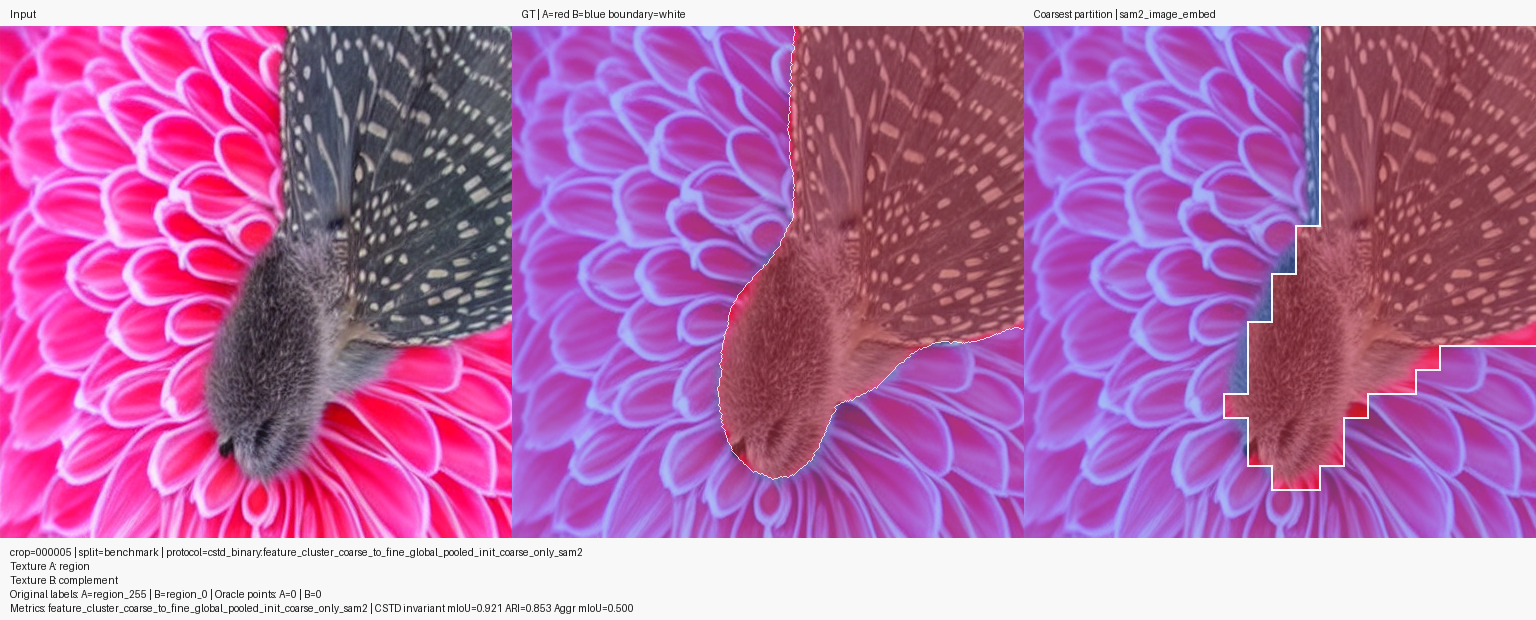} \\
        \includegraphics[width=0.55\textwidth]{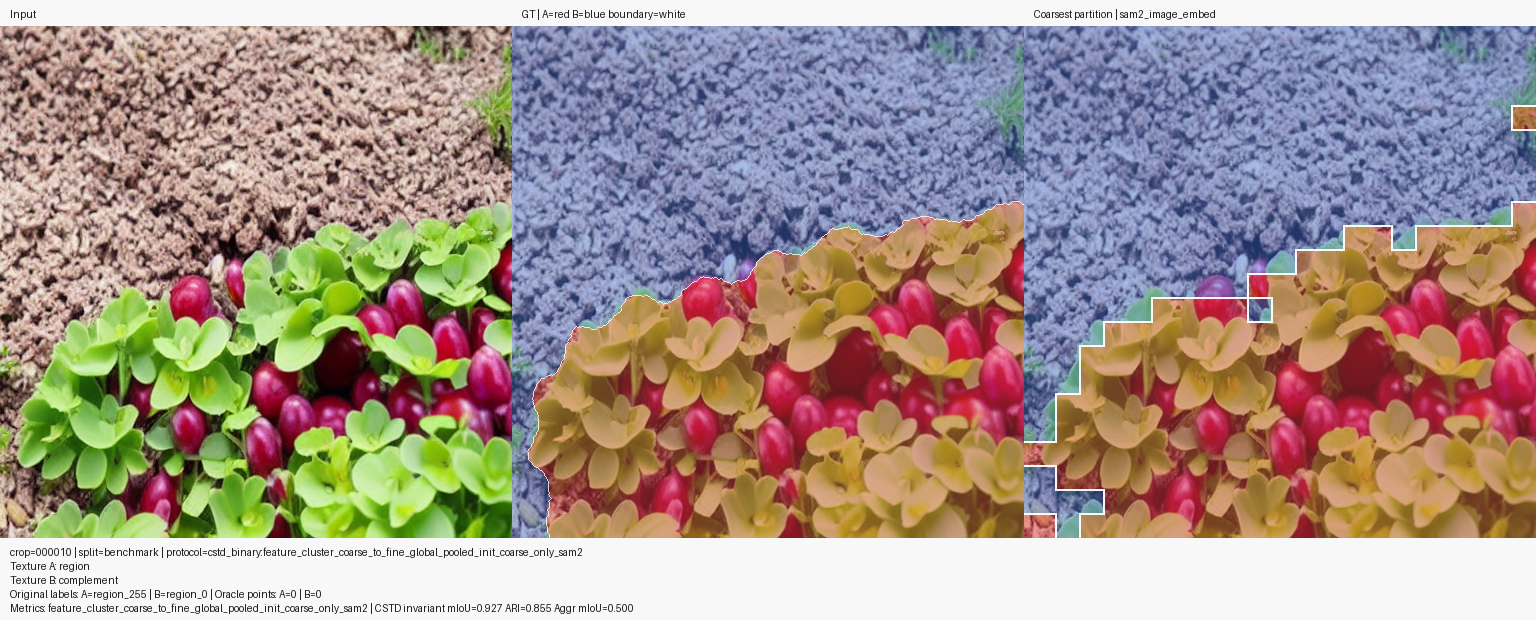} \\
        \includegraphics[width=0.55\textwidth]{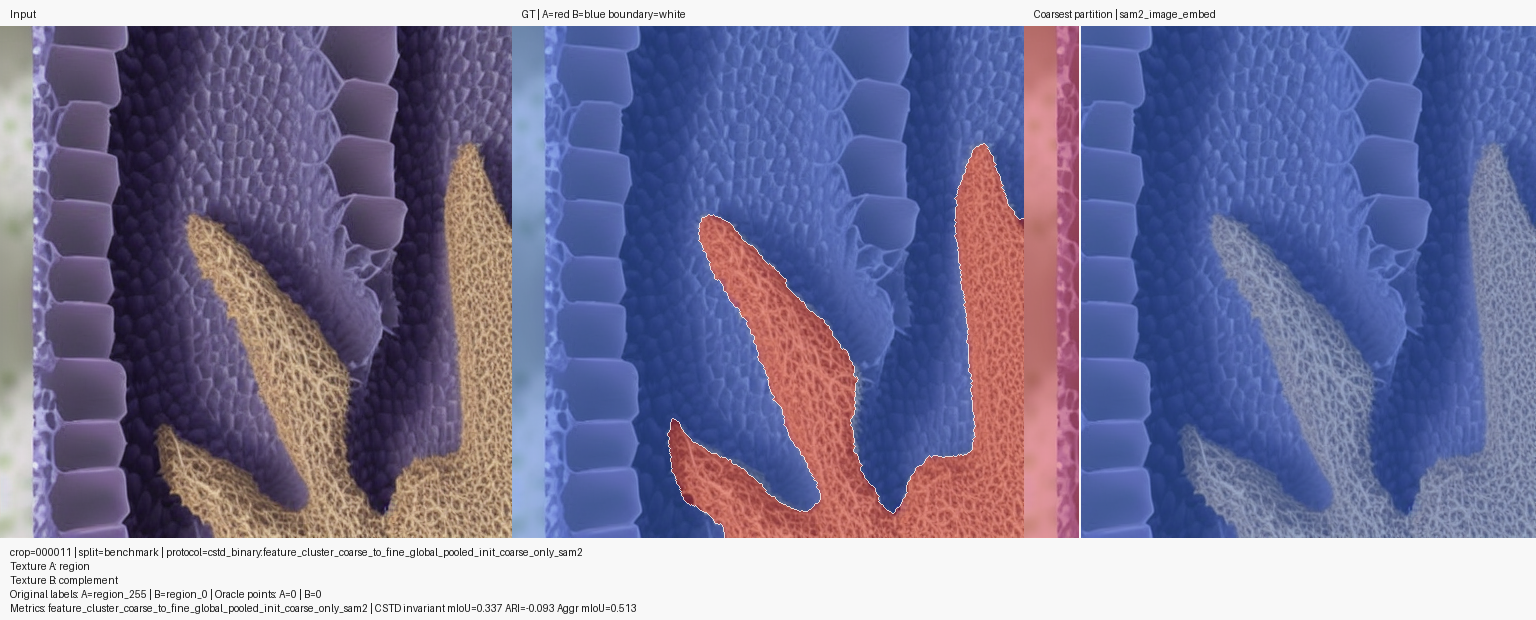} \\
        \includegraphics[width=0.55\textwidth]{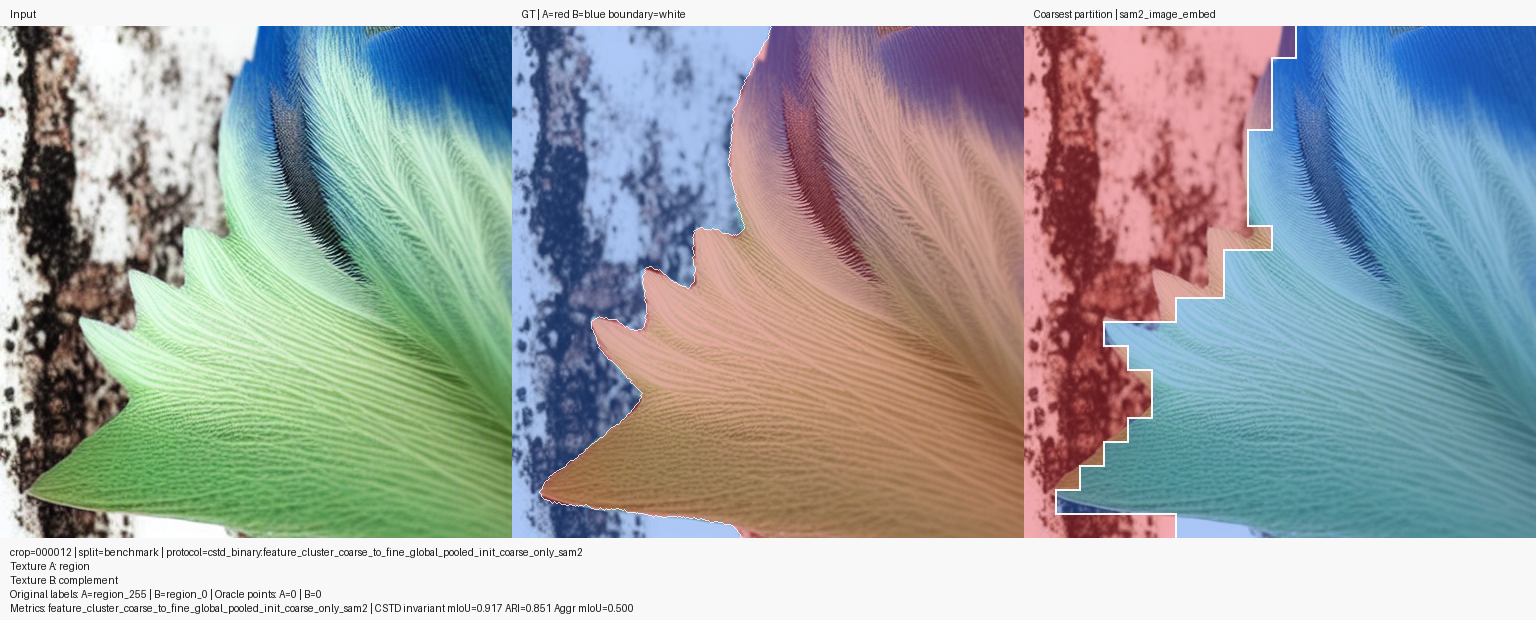} \\
    \end{tabular}
        \caption{ControlNet bridge: generated transition examples.}
    \label{fig:controlnet_bridge_app}
\end{figure*}

\begin{table*}[t]
\caption{ControlNet bridge retained evaluation artifact. The paper treats this route as a retained evaluation artifact with fixed retained images, predictions, and manifests under manifest-level audit. It is a complement to STLD rather than a replacement for STLD or natural benchmarks. The artifact contains 1,742 retained evaluation images and about 800 MB of packaged data. The main text emphasizes the invariant view; this appendix table retains the direct and matched-subset views together.}
\label{tab:controlnet_secondary}
\centering
\scriptsize
\begin{tabularx}{\textwidth}{@{}Xlcc@{}}
\toprule
Readout / comparator & Subset / coverage & Direct mIoU / ARI & Invariant mIoU / ARI \\
\midrule
TextureSAM public checkpoint rerun & all-1742, matched subset & 0.4212 / 0.6446 & 0.6425 / 0.5510 \\
Proposal-space recovery (Stage-A) & all-1742, matched subset & 0.4993 / 0.6039 & 0.6803 / 0.6039 \\
TextureSAM public checkpoint rerun & common-1739 & 0.4219 / 0.6457 & 0.6436 / 0.5519 \\
Proposal-space recovery (Stage-A) & common-1739 & 0.4993 / 0.6039 & 0.6803 / 0.6039 \\
\bottomrule
\end{tabularx}
\end{table*}

\begin{table}[t]
\caption{Prediction coverage for the four central routes.}
\label{tab:prediction_coverage}
\centering
\scriptsize
\setlength{\tabcolsep}{4pt}
\renewcommand{\arraystretch}{1.05}
\begin{tabular}{lcccc}
\toprule
Readout / comparator & RWTD & ADE20K-selected & STLD & ControlNet \\
\midrule
SAM2.1-small & 256/256 & 687/687 & 199/200 & 1742/1742 \\
TextureSAM checkpoint & 253/256 & 628/687 & 182/200 & 1739/1742 \\
Proposal-space recovery & 256/256 & 687/687 & 200/200 & 1742/1742 \\
\bottomrule
\end{tabular}
\end{table}

\section{Matched-Subset and Coverage Views}
\label{app:matched_subset_views}

\begin{table*}[t]
\caption{Matched-subset comparison (common predictions only).}
\label{tab:matched_subset_views}
\centering
\scriptsize
\setlength{\tabcolsep}{3pt}
\renewcommand{\arraystretch}{1.05}
\begin{tabularx}{\textwidth}{@{}llccc@{}}
\toprule
Route & Readout / comparator & Coverage & mIoU & ARI \\
\midrule
\multicolumn{5}{@{}l}{\textbf{RWTD -- official evaluator}} \\
TextureSAM public checkpoint rerun~\cite{texturesamrepo2026}
& common subset & 253/253 & 0.4684 & 0.6163 \\
Proposal-space recovery
& common subset & 253/253 & 0.4645 & \textbf{0.7013} \\
\midrule
\multicolumn{5}{@{}l}{\textbf{ADE20K-selected crops -- partition-invariant evaluator}} \\
TextureSAM public checkpoint rerun~\cite{texturesamrepo2026}
& common subset & 728/728 & \textbf{0.4963} & 0.3689 \\
Proposal-space recovery (Stage-A)
& common subset & 728/728 & 0.4855 & \textbf{0.3756} \\
\midrule
\multicolumn{5}{@{}l}{\textbf{STLD -- direct foreground evaluator}} \\
TextureSAM public checkpoint rerun~\cite{texturesamrepo2026}
& common subset & 182/182 & 0.5140 & 0.7526 \\
Proposal-space recovery
& common subset & 182/182 & \textbf{0.7195} & \textbf{0.7791} \\
\midrule
\multicolumn{5}{@{}l}{\textbf{ControlNet bridge -- partition-invariant evaluator}} \\
TextureSAM public checkpoint rerun~\cite{texturesamrepo2026}
& common subset & 1739/1739 & 0.6436 & 0.5519 \\
Proposal-space recovery (Stage-A)
& common subset & 1739/1739 & \textbf{0.6803} & \textbf{0.6039} \\
\bottomrule
\end{tabularx}
\end{table*}

\subsection{Optional Appendix-Only CAID Breadth Check}
\label{app:caid_secondary}

Remote-sensing imagery with shoreline partitions at larger spatial scale.

\begin{figure*}[t]
    \centering
    \includegraphics[width=0.98\textwidth]{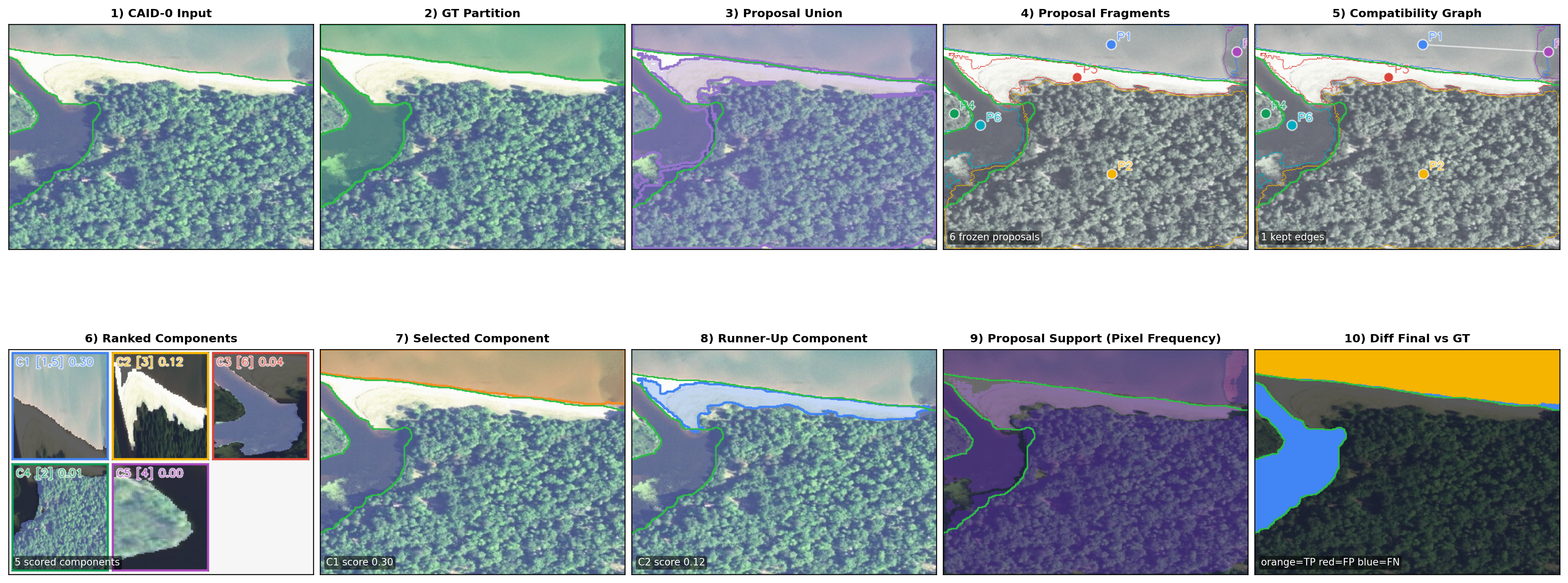}
    \caption{CAID shoreline breadth check.}
    \label{fig:caid_app}
\end{figure*}

\begin{table*}[t]
\caption{Optional appendix-only CAID breadth check under the partition-invariant evaluator. CAID is retained separately because its exact shoreline masks and large contiguous land/water partitions are useful but narrower than the manuscript's central natural/synthetic texture settings.}
\label{tab:caid_secondary}
\centering
\scriptsize
\begin{tabularx}{\textwidth}{@{}Xlc@{}}
\toprule
Readout / comparator & Subset / coverage & Invariant mIoU / ARI \\
\midrule
TextureSAM public checkpoint rerun & all-3104, 3063/3104 & 0.6691 / 0.5080 \\
Proposal-space recovery (Stage-A) & all-3104, 3104/3104 & 0.6677 / 0.5745 \\
TextureSAM public checkpoint rerun & common-3063 & 0.6780 / 0.5148 \\
Proposal-space recovery (Stage-A) & common-3063 & 0.6677 / 0.5745 \\
\bottomrule
\end{tabularx}
\end{table*}

\clearpage
\section{Audit and Settings}
\label{app:artifact_map}

\subsection{Artifact Boundaries}

\noindent\textbf{Feature-space route:} Reproducible at retained-summary level (parity tables, holdout previews, scale summaries).

\begin{table}[t]
\caption{Feature-space probe: retained artifacts.}
\label{tab:feature_artifact_map}
\small
\centering
\begin{tabular}{lp{0.6\textwidth}}
\toprule
Artifact & Purpose \\
\midrule
Parity summary & Reproduces Table~\ref{tab:feature_parity} \\
Holdout previews & Qualitative examples \\
Scale summary & Figure~\ref{fig:feature_scale_diag} \\
Assembly scripts & Regenerate figures \\
\bottomrule
\end{tabular}
\end{table}

\noindent\textbf{Complements:} Reproducible at manifest level (fixed crop/mask manifests, subset IDs).

\begin{table}[t]
\caption{Complement routes: retained artifacts.}
\label{tab:complement_artifact_map}
\small
\centering
\begin{tabular}{lp{0.6\textwidth}}
\toprule
Artifact & Purpose \\
\midrule
ADE20K manifest, masks & 56 crop tests \\
ControlNet scaffold, archive & 1,742 images \\
Prediction manifests & Coverage tracking \\
\bottomrule
\end{tabular}
\end{table}

\subsection{Experimental Settings}
\label{app:experimental_settings}

\begin{table}[t]
\caption{Feature-space probe settings.}
\label{tab:feature_probe_settings}
\small
\centering
\begin{tabular}{lp{0.65\textwidth}}
\toprule
Setting & Value \\
\midrule
Backbone & Frozen SAM (\texttt{fpn\_2}) \\
Pooling & 3$\times$3 average, stride 3 to 24$\times$24 \\
Clustering & 2-way spherical k-means \\
Distance & Cosine similarity \\
Eval data & RWTD, STLD, CAID \\
Visualization & PCA RGB, alpha 0.7 \\
Model selection & Coarsest scale \\
\bottomrule
\end{tabular}
\end{table}

\begin{table}[t]
\caption{Proposal-space consolidation settings.}
\label{tab:proposal_recovery_settings}
\small
\centering
\begin{tabular}{lp{0.65\textwidth}}
\toprule
Setting & Value \\
\midrule
Encoder & ConvNeXt-Tiny PTD (80K/20K images) \\
Descriptor & 384-D L2-normalized \\
Merge learner & HistGBT (lr=0.05, depth=10, 400 iter) \\
Core selection & Conservative low-overmerge \\
Rescue & RWTD-only classifier + regressor \\
Encoder optimizer & AdamW: lr=3e-4, 2 epochs, batch=48 \\
Thresholds & Stage-A: 0.44; others: 0.08--0.41 \\
Target tuning & No RWTD-label tuning \\
\bottomrule
\end{tabular}
\end{table}

\begin{table}[t]
\caption{ControlNet bridge generation settings.}
\label{tab:controlnet_bridge_settings}
\small
\centering
\begin{tabular}{lp{0.65\textwidth}}
\toprule
Setting & Value \\
\midrule
Base model & Stable Diffusion v1.5 \\
ControlNet & 500-step transition-tuned \\
Scaffold & Exact two-region stitch, inherited labels \\
Prompt & ``One transition between two texture patches'' \\
Denoising & 30 steps, CFG=7.5, control=1.0 \\
Noising & $t'=650$, strength=0.6 \\
Resolution & 512 px \\
Archive & 1,742 images, seed=42 \\
Augmentation & Perlin (prob=0.7) \\
\bottomrule
\end{tabular}
\end{table}

\subsection{Compute and Artifacts}
\label{app:compute_summary}

\begin{table}[t]
\caption{Compute and retained artifacts.}
\label{tab:compute_resources}
\small
\centering
\begin{tabular}{lp{0.65\textwidth}}
\toprule
Route & Status \\
\midrule
Feature-space probe & Retained summaries (no rerun) \\
RWTD proposal-space & 430 MiB Stage-A, 637 MiB full; 0.025--0.378 s/img \\
STLD proposal-space & Stage-A, 200 images \\
ADE20K-selected & Stage-A, 687 crops \\
ControlNet bridge & Stage-A, 1,742 images \\
TextureSAM reruns & Baseline manifests \\
SAM2.1-small reruns & Comparator manifests \\
ControlNet generation & 1,742 retained images \\
\bottomrule
\end{tabular}
\end{table}

\subsection{Asset Licenses}
\label{app:asset_licenses}

\begin{table}[t]
\caption{Reused assets and licenses.}
\label{tab:asset_licenses}
\small
\centering
\begin{tabular}{lp{0.25\textwidth}p{0.25\textwidth}p{0.25\textwidth}}
\toprule
Asset & Type & License & Usage \\
\midrule
SAM & model/code & Apache 2.0 & Mask generator \\
SAM2 & model/code & Apache 2.0 & Comparator \\
TextureSAM & checkpoint & Public & Baseline \\
DTD & dataset & Research & Textures \\
PTD & dataset & Unconfirmed & Supervision \\
RWTD & dataset & Unconfirmed & Eval \\
STLD/Brodatz & lineage & Unconfirmed & Supervision \\
ADE20K & dataset & ToU & Eval \\
ControlNet & model/code & Apache 2.0 & Bridge control \\
Stable Diffusion & model & RAIL++-M & Generation \\
CAID & dataset & Unconfirmed & Breadth \\
ControlNet-500 & checkpoint & Research only & Bridge gen \\
\bottomrule
\end{tabular}
\end{table}

\end{document}